\documentclass[journal,comsoc]{IEEEtran}
\usepackage{cite}

\ifCLASSINFOpdf
\else
\fi
\usepackage{array}
\usepackage{url}
\usepackage{graphicx}
\usepackage{amsmath}
\usepackage{multirow}
\usepackage{makecell}
\usepackage{hyperref}
\usepackage{algorithm}
\usepackage{algpseudocode}
\usepackage[table]{xcolor}

\usepackage{orcidlink}

\usepackage{booktabs}  
\usepackage{siunitx}   

\begin{document}
%
\title{Progressive Multi-Source Domain Adaptation for Personalized Facial Expression Recognition}
%
%
%

\author{Muhammad~Osama~Zeeshan \orcidlink{0009-0006-1463-2465},~\IEEEmembership{Student Member,~IEEE,}, Marco~Pedersoli \orcidlink{0000-0002-7601-8640
},~\IEEEmembership{Member,~IEEE,} Alessandro Lameiras Koerich \orcidlink{0000-0001-5879-7014},~\IEEEmembership{Member,~IEEE,}
        and~Eric~Granger \orcidlink{0000-0001-6116-7945} , ~\IEEEmembership{Member,~IEEE}
\thanks{The authors are affiliated with the LIVIA and ILLS, the Department of Systems Engineering, and the Department of Software Engineering at ETS Montreal, Canada.}
}

%
%

\markboth{Zeeshan et al. [Transactions on Affective Computing 2025]}%
{Shell \MakeLowercase{\textit{Zeeshan et al.}}: Bare Demo of IEEEtran.cls for IEEE Journals}


%



\maketitle

\makeatletter
\newcommand{\thickhline}{%
    \noalign {\ifnum 0=`}\fi \hrule height 1pt
    \futurelet \reserved@a \@xhline
}


\begin{abstract}
Personalized facial expression recognition (FER) involves adapting a machine learning model using samples from labeled sources and unlabeled target domains. Given the challenges of recognizing subtle expressions with considerable interpersonal variability,
state-of-the-art unsupervised domain adaptation (UDA) methods focus on the multi-source UDA (MSDA) setting, where each domain corresponds to a specific subject, and improve model accuracy and robustness.
However, when adapting to a specific target, the diverse nature of multiple source domains translates to a large shift between source and target data. 
State-of-the-art MSDA methods for FER address this domain shift 
by considering all the sources to adapt to the target representations. 
Nevertheless, adapting to a target subject presents significant challenges due to large distributional differences between source and target domains, often resulting in negative transfer. 
In addition, integrating all sources simultaneously increases computational costs and causes misalignment with the target.
To address these issues, we propose a progressive MSDA approach that gradually introduces information from subjects (source domains) based on their similarity to the target subject. 
This will ensure that only the most relevant sources from the target are selected, which helps avoid the negative transfer caused by dissimilar sources. 
During adaptation, the source domains are introduced in a curriculum manner. We first exploit the closest sources to reduce the distribution shift with the target and then move towards the furthest while only considering the most relevant sources based on the predetermined threshold.
%
Furthermore, to mitigate catastrophic forgetting caused by the incremental introduction of source subjects, we implemented a density-based memory mechanism that preserves the most relevant historical source samples for adaptation. 
Our extensive experiments \footnote[1]{ \url{https://github.com/osamazeeshan/P-MSDA}} show the effectiveness of our proposed method on challenging FER datasets: Biovid, UNBC-McMaster, Aff-Wild2, and BAH. Further, performance is evaluated on a cross-dataset setting \emph{(UNBC-McMaster $\rightarrow$ BioVid)}, showing the importance of gradually adapting to source subjects.
\end{abstract} 


%
\IEEEpeerreviewmaketitle

\section{Introduction}
\label{sec:intro}
\IEEEPARstart{I}{n} recent years, there has been a growing demand for deep learning (DL) models that can perform well on FER across various industrial sectors such as in detecting suspicious or criminal behavior, automated emotion recognition, or the estimation of pain in health care  \cite{aslam2023privileged, sharafi25, aslam2024distilling, waligora2024joint}. Addressing the significant variability of facial expressions between individuals due to cultural and ethnic differences or varying capture conditions is a challenging problem because of the subtle expression in real-world applications that leads to a substantial disparity between the data used to train and test DL models \cite{ben2010theory}. Therefore, adapting a deep FER model to a specific individual (i.e., personalization) is important to maintain a high level of performance. 

Personalized FER has been extensively studied in the literature, primarily through 
supervised learning approaches and fine-tuning techniques \cite{rescigno2020personalized, barros2019personalized, zen2016learning} to capture individual-specific nuances. These approaches mostly rely on fully or weakly labeled data to adapt and create a personalized model for each subject.
Fine-tuning a model using fully labeled data requires a costly annotation of samples, and this annotation may be ambiguous due to variability among the annotators \cite{ren2015faster, hu2018squeeze}.
Unsupervised domain adaptation (UDA) \cite{gretton2012kernel, long2016unsupervised} is a promising alternative for leveraging unlabeled data in FER. Nevertheless, SOTA UDA techniques treat datasets as domains containing samples from mixed subjects across source and target domains, which limits the model's ability to perform fine-grained adaptation \cite{han2020personalized, li2018deep, zhu2016discriminative}.
Defining each subject as a domain can address this issue yet blending source data to perform UDA restricts its capacity to handle variations and diversity within the target domain. 
To address this challenge, multi-source (unsupervised) domain adaptation (MSDA) \cite{zhao2021madan, zhao2020multi, zeeshan2024subject} has gained significant popularity as it incorporates information from multiple source domains to enhance model resilience to different target variations. 
In \cite{zeeshan2024subject}, the authors introduced a subject-based domain adaptation method, where each domain consists of a distinct subject, thus multiple sources and a single target subject. The target domain consists of samples belonging to a single person, where the data are captured in a stationary environment and incorporate less effective diversity. This approach differs from the traditional MSDA methods that adapt to a target domain of mixed individuals. The subject-based method focused on adapting to a single individual enables more precise and targeted adaptation strategies to create a personalized FER model.  


\begin{figure*}[t!]
\centering
\includegraphics[ width=0.95\linewidth]{ 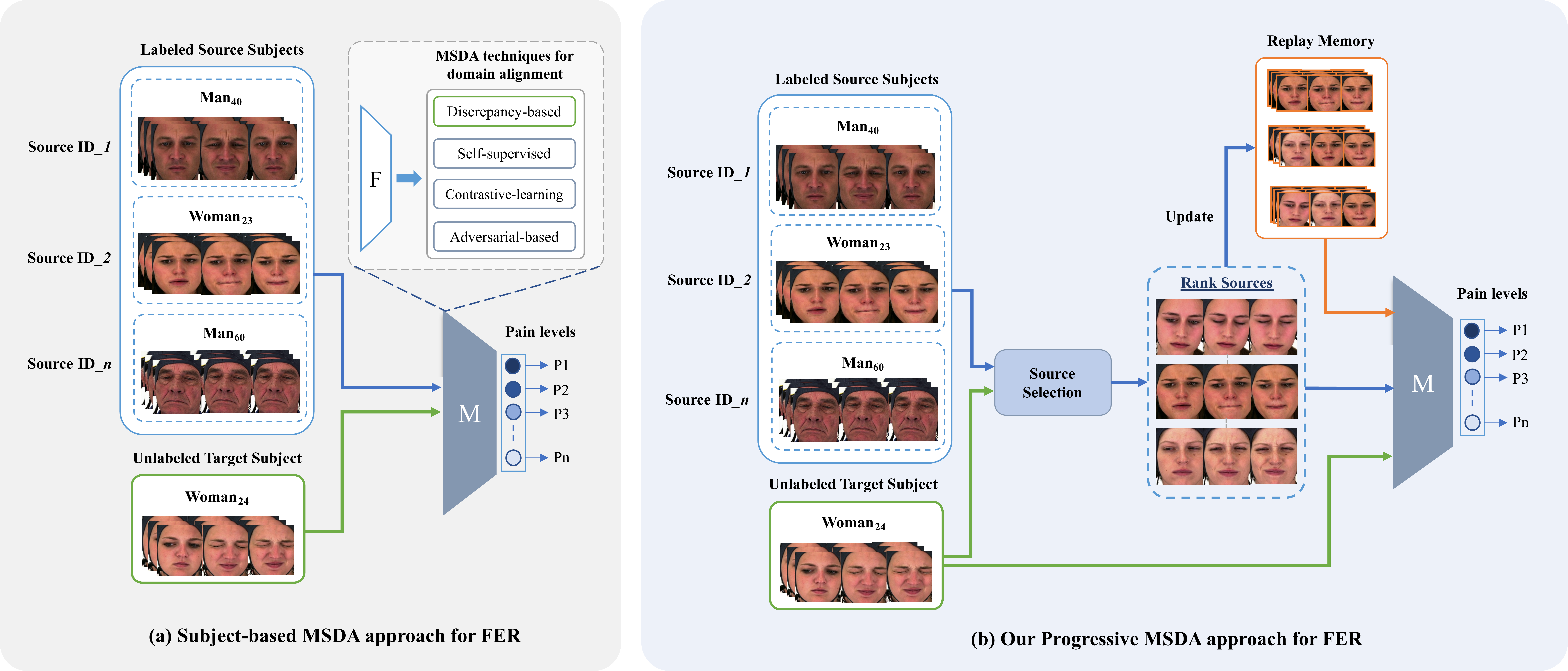}
\caption{Comparison between subject-based MSDA and our proposed progressive MSDA. (a) Subject-based MSDA aligns all source domains simultaneously with the target using a \emph{Discrepancy-based, Self-supervised, Contrastive-learning, or Adversarial-based} approaches. (b) Our Progressive MSDA first rank and gradually adapts source domains (subjects) based on their similarity to the target domain,
optimizing the transfer process through sequential adaptation. Secondly, we construct a replay memory (domain) that retains key samples from previously adapted source domains, which are re-accessed after each adaptation.  The discrepancy-based approach is applied to align the source and target domains.}

\label{fig:motivate}      
\vspace{-10pt}
\end{figure*}
MSDA methods address the challenge of transferring knowledge across diverse domains (datasets) by aligning multiple source domains with a target domain representation. By leveraging data from multiple sources, MSDA aims to enhance model robustness and generalization by minimizing domain shift \cite{kang2020contrastive, scalbert2021multi} and create a shared feature space where source and target distributions are closely matched. While effective in some cases, this approach forces all sources to align with the target data, often overlooking individual domain variability. In contrast, the subject-based domain adaptation method in \cite{zeeshan2024subject} struggles with target adaptability under such alignment, as the inherent differences among source subjects can deviate significantly from the target domain.
Although the adaptation to all the source subjects is beneficial for learning a better representation of diverse subjects, adapting to a single target subject does not always require incorporating every source subject. This is due to the subject's negative transfer \cite{rosenstein2005transfer, wang2019characterizing} caused by the large difference in distributions of source and target domains. For instance, if the target domain corresponds to a \emph{young caucasian woman} and the source domains consist of a diverse group of individuals from varying personal characteristics such as age, ethnicity, and gender (illustrated in the Fig.~\ref{fig:motivate}(a)). Integrating all the source domains will generate a large domain shift between source and target, causing a model to deviate, and creating difficulties in the adaptation process. Furthermore, it would not be feasible in real-world applications due to high computing and memory consumption.  

To address the challenges of developing a subject-based MSDA method for FER, we gradually introduce only the most relevant source domains (subjects) during the adaptation process.
We introduce a method that integrates curriculum learning (CL) \cite{bengio2009curriculum} and self-paced learning (SPL) \cite{kumar2010self}, leveraging prior knowledge from CL to prioritize easier source subjects during target domain adaptation. Simultaneously, SPL is employed to dynamically adapt the learning process, allowing the model to recalibrate and select relevant sources as training progresses. To effectively capture contextual information from each source subject while avoiding domain corruption, we adopt a progressive approach by adapting to one source at a time gradually. 
Drawing inspiration from domain-incremental learning (DIL) replay techniques \cite{de2019continual, parisi2019continual}, where new conditions are incrementally introduced to the target domain while preserving previously learned information. In this work, we propose a replay-based mechanism that dynamically selects and stores the most transferable samples from the visited source domains. This ensures efficient adaptation to the target subject prevents catastrophic forgetting, and reduces computational overhead.

In this paper, we introduce a new paradigm of progressive multi-source domain adaptation (P-MSDA) for personalized facial expression recognition (FER), where the goal is to adapt to a single unlabeled target subject by \emph{progressively} leveraging the most relevant source subjects. Unlike conventional UDA or MSDA methods that adapt at a generic domain level or aggregate all sources at once, our approach focuses on subject-level personalization, recognizing that the distribution of facial expression can vary significantly across individuals due to factors such as demographics, age, and cultural background. P-MSDA begins by ranking source subjects based on their similarity to the target in feature space and selecting only those above a threshold, ensuring that adaptation starts from the most relevant sources as illustrated in Fig.~\ref{fig:motivate}(b). As training progresses, we re-calculate similarities to dynamically incorporate new sources, thereby combining the principles of curriculum learning \cite{bengio2009curriculum} (starting from “easier” sources) and self-paced learning (adapting as the model matures).

To train such a model, a naive approach begins with the easiest source subject and gradually moves towards the hardest subjects while storing all the previously seen subjects. Another option is to train a model by ignoring the previously learned subjects and only training with the newly added source subject. The former approach requires significant computation power and memory consumption, while the latter setting will lean towards the problem of catastrophic forgetting \cite{aljundi2017expert,laurensi2024alleviating}. 
To avoid using every visited subject and eliminate the problem of forgetting, we follow the DIL replay-based \cite{isele2018selective} strategy, where we create a replay dictionary that preserves a small set of previously adapted samples based on the closest source points from the target clusters and only incorporates those with the newly added subject. We keep updating the replay dictionary as the training progresses. Our method aims to strike a balance between leveraging diverse source information and maintaining the domain-specific characteristics of the target.

The main contributions of this paper are summarized as follows. \textbf{(1)} A novel progressive learning framework for MSDA personalized FER that exploits the closest source subjects for target adaptation. Additionally, we present an effective training strategy based on DIL that starts with the closest source and gradually introduces a new subject while limiting the number of source subjects to the \emph{top-N} pertinent sources. This approach minimizes the risk of data corruption from mixed source domains and reduces computational complexity.
\noindent \textbf{(2)} Inspired by the incremental learning replay technique, we present a density-based sample selection mechanism that retains the most relevant samples from previously visited source subjects, avoiding catastrophic forgetting.
\noindent \textbf{(3)} The performance of the proposed method is extensively evaluated on diverse groups of source and target subjects using two benchmark pain estimation datasets, BioVid and UNBC-McMaster, as well as two in-the-wild affective computing datasets, Aff-Wild2 and Behavioral Ambivalence/Hesitancy (BAH). We also show the efficacy of our method by evaluating cross-dataset, specifically UNBC-McMaster (source) to BioVid (target). Further, we present a comprehensive analysis of the selection of previous samples for better target adaptation.


\section{Related Works}
\label{sec:relwork}

\subsection{Personalized Facial Expression Recognition} 

Personalization in facial expression recognition (FER) focuses on adapting models for specific individuals by considering variations in facial features, cultural nuances, and subjective labeling. In recent years, supervised personalization techniques \cite{ barros2019personalized, rescigno2020personalized} have gained significant attention, as they aim to enhance model performance using labeled data, which leads to improved accuracy. Despite the progress made, these methods depend on fine-tuning a model using subject-specific labels, which are often unavailable in real-world applications. Another challenge arises from users with extreme facial variations, which can lead to issues with identity bias and temporal dynamics, particularly when dealing with unseen subjects. To tackle these issues, domain adaptation (DA) techniques ~\cite{zhu2016discriminative, han2020personalized} have been explored. These methods align the feature distributions of labeled (seen) data with subject-specific (unseen) data that is not explicitly labeled.

\subsection{Domain Adaptation}  
Domain adaptation (DA) methods adapt from labeled source domains to unlabeled target domains, and can be broadly grouped into discrepancy-based~\cite{zhu2016discriminative, han2020personalized}, reconstruction-based~\cite{ghifary2016deep, zhu2017unpaired}, and adversarial-based~\cite{chen2021cross, odena2017conditional, yan2018unsupervised} approaches. While effective, most DA methods rely on a single source domain, limiting their ability to generalize across diverse data distributions. To overcome this, our prior work on subject-based MSDA~\cite{zeeshan2024subject} introduced multiple source domains (subjects), showing improved performance by leveraging inter-subject diversity.
\subsection{Multi-Source Domain Adaptation}
Multi-source domain adaptation (MSDA) leverages multiple labeled source datasets to improve target domain accuracy and robustness. Existing MSDA approaches in image classification~\cite{kang2020contrastive, peng2019moment, nguyen2021stem, zhao2021madan} include adversarial~\cite{zhao2021madan, nguyen2021stem}, self-supervised~\cite{venkat2020your, deng2022robust}, and discrepancy-based~\cite{kang2020contrastive} strategies, which mitigate domain shift effectively on standard benchmarks~\cite{peng2019moment}.
However, their application to subject-based facial expression recognition (FER) remains limited, especially when scaling to many source domains. Zeeshan et al.~\cite{zeeshan2024subject} addressed this by adapting to a target subject from up to 77 sources, selecting the top-$k$ most similar sources for joint training. While effective, aggregating multiple sources at once can introduce noise from less relevant domains.
In contrast, our method ranks sources by distributional similarity to the target and incorporates them progressively, allowing fine-grained adaptation while mitigating negative transfer from less relevant subjects.
\subsection{Gradual Domain Adaptation}
Our work is also closely related to gradual domain adaptation (GDA). In GDA, including intermediate domains helps reduce the significant domain shift between source and target. The source domain adapts to these intermediate domains, gradually bridging the gap between the target domain. Several methods are introduced that help in gradual adaptation, such as using self-training \cite{zhou2022active, chen2021gradual, wang2022understanding} where intermediate domains were defined, in the absence of intermediate domains \cite{abnar2021gradual}. Our approach, inspired by GDA, directly uses the source domains that are closer to the target in the feature space and uses them for the target adaptation instead of going from source to target with the help of intermediate domains. Nevertheless, based on the model selection criteria, we only incorporate source domains that benefit the target.
\subsection{Incremental Learning}
Incremental learning (IL) sequentially acquires knowledge from multiple datasets without retaining full access to past data and is generally categorized into task-, class-, and domain-incremental learning (DIL)\cite{hsu2018re}. This work focuses on DIL, which adapts to new domains while avoiding catastrophic forgetting.
DIL methods have been applied in image classification\cite{de2019continual}, autonomous driving~\cite{mirza2022efficient}, and semantic segmentation~\cite{michieli2019incremental}, and typically fall into three categories: parameter isolation~\cite{xu2018reinforced}, regularization-based approaches (e.g., knowledge distillation~\cite{hou2019learning, zhang2020class}), and replay-based strategies~\cite{isele2018selective, laurensi2024alleviating, shin2017continual}. For domain adaptation, prior works~\cite{wei2020incremental, kiran2022incremental} address multiple target domains incrementally, but often overlook domain shift.
In contrast, our method adapts to a single unlabeled target by progressively introducing source domains while selectively preserving only the most relevant samples, thereby mitigating domain discrepancies and improving target personalization.

\subsection{Self-paced Curriculum Learning}

{Curriculum learning (CL)~\cite{bengio2009curriculum} and self-paced learning (SPL)~\cite{kumar2010self} start with easier tasks or samples and progressively move to more complex ones. CL leverages prior knowledge to guide the sequence, while SPL adjusts dynamically to the learner’s pace. Several works adopt these ideas for UDA, e.g., Choi et al.~\cite{choi2019pseudo} select high-density target samples first, and Wang et al.~\cite{wang2022self} choose target samples agreed upon by multiple source classifiers. Jiang et al.~\cite{jiang2015self} combine both paradigms into a unified self-paced curriculum learning (SPCL) framework. Yang et al.~\cite{yang2020curriculum} apply CL to MSDA by selecting source samples similar to the target distribution. However, these approaches typically focus on confident target samples, without considering how source samples are introduced. More recently, Zhu et al.~\cite{zhu2025tensorial} introduced Tensorial Multiview Low-Rank High-Order Graph Learning (MLRGL), which progressively refines pseudo-labels via masked–unmasked consistency and propagates multiview features through a high-order graph. While conceptually related to our progressive pseudo-labeling, MLRGL targets generic multiview UDA, whereas our method addresses a problem of personalized multi-source adaptation where both subject similarity and progressive inclusion of source subjects are crucial.
}

\begin{figure*}[t!]
\centering
\includegraphics[width=1.0\linewidth]{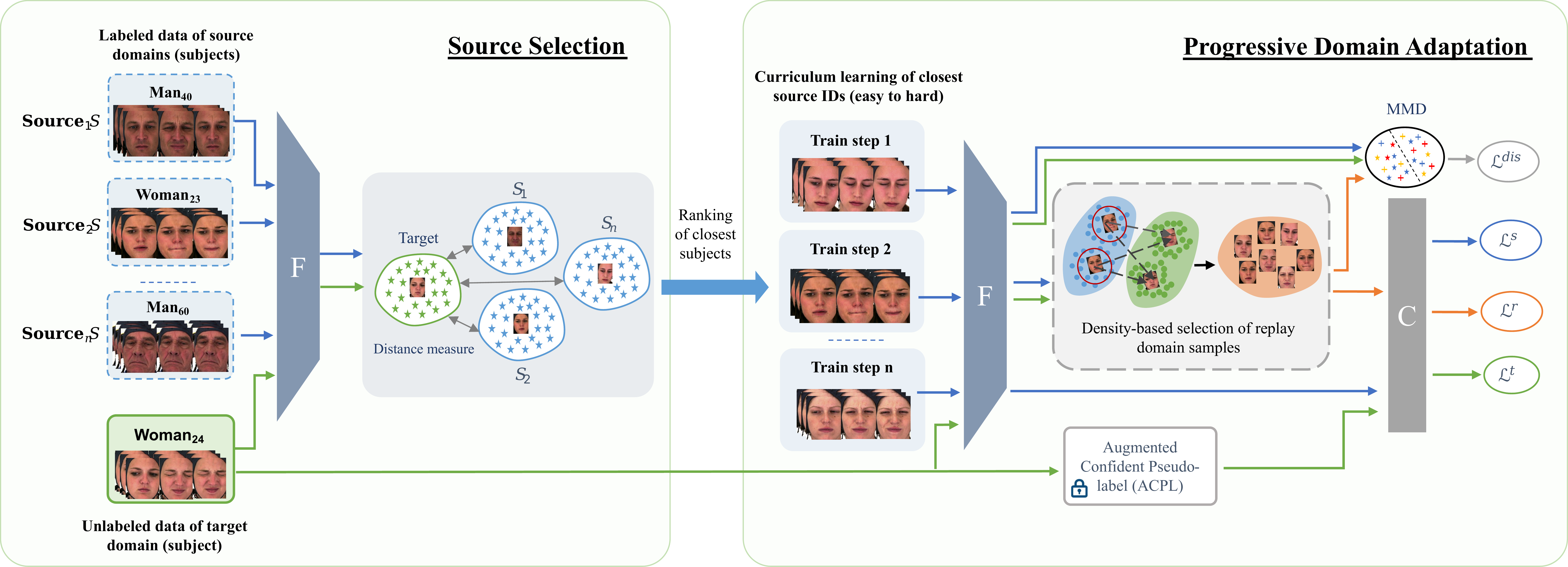}
\caption{Overview of our proposed progressive MSDA method for the adaptation to the target subject. \textbf{Source Selection Phase:} We estimate the similarity matrix between every source and target embedding, followed by ranking the sources from most to least similar subjects. \textbf{Progressive Domain Adaptation Phase:} 
Ranked sources are progressively incorporated through iterative training steps (\textit{Train Step-1}, \textit{Train Step-2}, ..., \textit{Train Step-n}). At each step, a new source subject is introduced and aligned with the target by calculating discrepancy and supervised losses. The Augmented Confident Pseudo-label (ACPL) technique from \cite{zeeshan2024subject} generates reliable pseudo-labels for the target. Finally, we create a replay dictionary using a density-based selection to preserve previously visited relevant source samples.}
\label{fig:dimsa_network}      
\vspace{-10pt}
\end{figure*}




\section{Proposed Approach}
\label{sec:proposed}
Fig.~\ref{fig:dimsa_network} provides an overview of the P-MSDA framework, which dynamically generates a curriculum of source subjects, ranging from the easiest to the hardest, gradually introduced to a given target domain (Sec.~\ref{sec:selc_src_sbs}), followed up by the creation of a dynamic replay dictionary (domain) of the most representative source distributions that helps in preventing catastrophic forgetting during the adaptation process (Sec.~\ref{sec:selc_relv_samp}).

\subsection{Preliminaries}





In MSDA, given a set of labeled source domains $\mathcal{S}={\{\mathbf{S}_1,\mathbf{S}_2,\ldots,\mathbf{S}_a,\ldots,\mathbf{S}_D \}}$ and a single unlabeled target domain $\mathbf{T}$, where $a = {\{1,2,\ldots,D}\}$ is the number of source domains. To preserve relevant samples from previous source domains, we define a replay domain $\mathbf{R}$. We define a source domain as $\mathbf{S}_a = {\{(\mathbf{x}_i^\text{s}, y_i^\text{s})}\}_{i=1}^{N^\text{s}}$, where  $N^\text{s}$ represents the number of samples within each source domain $a$. Assuming that $\mathbf{x}_i^\text{s}$ represents an input embedding of a source domain sample $i$ produced by an encoder $F$, the labeled source domain input space is defined as $\mathbf{x}_i^\text{s}$, and their respective labels as $y_i^\text{s}$. We define a single unlabeled target domain as $\mathbf{T} = {\{\mathbf{x}_i^\text{t}\}}_{i=1}^{N^\text{t}}$, where $N^\text{t}$ denotes the number of samples within the target domain. A replay domain is defined as $\mathbf{R}={\{(\mathbf{x}_i^\text{r}, y_i^\text{r})}\}_{i=1}^{N^\text{r}}$, where $N^\text{r}$ represents the number of relevant source domains from $\mathcal{S}$, which keeps on updating after adapting \textcolor{violet}{the model $\Theta$} to the target domain $\mathbf{T}$. The model $\Theta$ integrates the representation learning module $F$ with the classifier $C$.  The objective is to leverage the information of source domains by gradually introducing \textcolor{blue}{domains from} $\mathcal{S}$ to improve the performance of a model $\Theta$ on the target domain $\mathbf{T}$. Therefore, at each training step, the model $\Theta$ aims to learn from a new source domain from $\mathcal{S}$ and $\mathbf{R}$, which retain the most representative information from the previous source domains.

\subsection{Source Selection}
\label{sec:selc_src_sbs}

Before progressive adaptation of the model $\Theta$ to the target domain, we aim to select the most suitable domains from a multi-source domain $\mathcal{S}$ to align them with the target domain $\mathbf{T}$, based on a curriculum-based approach. Initially, we prioritize the source domains with high transferability to align them with the target domain. This will help select source domains with feature distributions similar to the target domain. After aligning the feature distributions of these source domains, a source selection\footnote{For algorithm, see Section I-A of the supplementary material} will prioritize the next round of source domains for alignment. As adaptation (training) continues, the model $\Theta$ gradually learn to focus on various aspects of the feature distribution to improve transferability. Our approach involves learning a curriculum to prioritize different source domains.
We hypothesize that source domains closest to the target domain in the feature space are the ones that are easier for the model to adapt to. We compute the cosine similarity between every source and the target domain in a mini-batch \cite{mondal2021mini}, domains are represented by their feature embeddings as:

\begin{equation}
    h(\mathbf{x}^\text{s}, \mathbf{x}^\text{t}) = \frac{\mathbf{x}^\text{s} \cdot \mathbf{x}^\text{t}}{\|\mathbf{x}^\text{s}\|_2 \cdot \|\mathbf{x}^\text{t}\|_2}
\end{equation}

\noindent where $\|.\|_2$ is the $\ell^2$-norm of the feature embeddings. 

Assuming a mini-batch of size $B$, where ${X}^\text{s}=\left\{\mathbf{x}^\text{s}_1, \mathbf{x}^\text{s}_2, \ldots, \mathbf{x}^\text{s}_B\right\}$ and ${X}^\text{t} =\left\{\mathbf{x}^\text{t}_1, \mathbf{x}^\text{t}_2, \ldots, \mathbf{x}^\text{t}_B\right\}$ represent feature embeddings from the source and target domains in the mini-batch, and $N^\text{b}$ is the total number of batches, a pair-wise similarity matrix is calculated as:
\begin{equation}
    \mathbf{M}^\text{c}(\mathcal{S}, \mathbf{T}) = \frac{1}{N^\text{b}}\sum_{i=1}^{N^\text{b}}{h}(X^\text{s}_i, X^\text{t}_i)
\end{equation}

To estimate the similarity matrix for all the source and target domain pairs, we define:
\begin{equation}
\label{eq:src_sel_cos}
\begin{aligned}
    P =[\mathbf{M}^\text{c}(\mathbf{S}_1, \mathbf{T}),\ldots, \mathbf{M}^\text{c}(\mathbf{S}_D, \mathbf{T})]
\end{aligned}
\end{equation}



\noindent where $\mathbf{M}^\text{c}(.)$ is the cosine similarity between every feature embedding in $\mathcal{S}$ and $\mathbf{T}$, $D$ represents the total number of source domains, and $P$ is a dictionary (list) that stores all pair-wise distances indexed by the respective source domain information. To determine the stopping criteria for including source domains {\color{blue} in the adaptation process}, we apply a normalization procedure on similarity measures in $P$ to scale them to the range $[0,1]$ and we obtain $\widetilde{P}$, which is the scaled version of $P$. Subsequently, we apply a predetermined threshold $\gamma$ to $\widetilde{P}$ \textcolor{blue}{to limit the number of source domains}. We formulate the process of selecting source domains as:
\begin{equation}
\label{eq:close_src}
\begin{aligned}
    \widetilde{\mathcal{S}}=\{\mathcal{S}_j: \widetilde{P}_j > \gamma \}  \quad \forall j \in \left\{ 1,\dots ,D\right\}
\end{aligned}
\end{equation}
\noindent {\color{blue} where $ \widetilde{\mathcal{S}}$ denotes the subset of source domains that meets the criterion $(\widetilde{P}_j > \gamma)$ and as a consequence, it is selected to update the model $\Theta$}. For each subset $\mathcal{\widetilde{S}}$ satisfying this condition, we compute a supervised loss as:

\begin{equation}
\label{eq:ce_src}
\begin{aligned}
\mathcal{L}^s=-\frac{1}{\widetilde{D}}\frac{1}{N_s}\sum_{d=1}^{\widetilde{D}}\sum_{i=1}^{N_s}{y}_i^d \cdot \log(C({\mathbf{x}}^d_i))
\end{aligned}
\end{equation}
\noindent where $\widetilde{D}$ indicates the selected source subject domains.
\noindent Note: After adapting $\mathbf{T}$ to the source domains included in ${\widetilde{\mathcal{S}}}$, we update the pair-wise distances using Eq.~\ref{eq:src_sel_cos} for the remaining source domains until we adapt the model $\Theta$ to the $top_s$ closest source domains to the target domain.






\subsection{Progressive Domain Adaptation of Target Subject}
\label{sec:selc_relv_samp}
Following the P-MSDA paradigm of aligning a given domain, we have access to a sorted $\widetilde{\mathcal{S}}$ comprised of source domains that are easy for the model $\Theta$ to align with the target. We gradually introduce source domains as $(\widetilde{\mathbf{S}}_1, \mathbf{R}, \mathbf{T};\Theta),(\widetilde{\mathbf{S}}_2, \mathbf{R}, \mathbf{T};\Theta),\ldots,(\widetilde{\mathbf{S}}_D, \mathbf{R}, \mathbf{T};\Theta)$. where $\Theta$ is the model that is updated at each step.

\noindent\textbf{Density-based Selection of Samples in Replay Domain.}
We preserve the relevant samples from the adapted source subject to avoid the forgetting issue while introducing a new source domain into the adaptation process. To select samples for the replay domain, we create source $\widetilde{\mathbf{S}}_a$ and target $\mathbf{T}$ clusters using density-based spatial clustering of applications with noise (DBSCAN) \cite{ester1996density}. The samples are selected based on the closely related points of the source cluster $\mathbf{K}^\text{s}$ to the target cluster $\mathbf{K}^\text{t}$.
We first estimate the local dense region of $\mathbf{K}^\text{s}$ by creating a matrix $\mathbf{H}^\text{s} \in \mathbb{F}^{K \times N^\text{s}}$, where $K$ is the number of clusters, and $N^\text{s}$ is the number of samples. Next, we calculate centroids $\mathbf{C}^\text{s} = \{\mathbf{c}^\text{s}_1, \mathbf{c}^\text{s}_2, \dots, \mathbf{c}^\text{s}_K\}$ for each cluster, and estimate the Euclidean distance between each sample and centroid as:

\begin{equation}
\label{eq:src_dis}
\begin{aligned}
   \mathbf{H}^\text{s}_{j,i} = \left\| \mathbf{x}_i - \mathbf{c}_j \right\|^2 \quad \forall j \in \{1, \dots, K\} ,  i \in \{1, \dots, N^\text{s}\}
\end{aligned}
\end{equation}
\noindent where $\mathbf{H}^\text{s}_{j, i}$ consists of multiple distances computed from  every $j$-th cluster. To pick the closest distance with the cluster centroids, we define:  

\begin{equation}
\label{eq:min_src_dis}
\begin{aligned}
   Z^\text{s}=\min_{j \in \{1, \dots, K\}} \{\mathbf{H}^\text{s}_{j,i} \}\quad \forall i \in \{1, \dots, N^\text{s}\}  
\end{aligned}
\end{equation}

\noindent where $Z^\text{s}$ is a list of distances $\{{z}_1, {z}_2, \dots, {z}_{N^\text{s}}\}$ sorted in ascending order 
to determine the closest samples to the cluster centroid. We sort the samples in $\mathbf{\widetilde{S}}_a$ based on the distances in ${Z^\text{s}}$. Subsequently, to determine the distances of $\mathbf{\widetilde{S}}_a$ from $\mathbf{T}$, we create target domain clusters $\mathbf{K}^\text{t}$ with centroids $\mathbf{C}^\text{t} = \{\mathbf{c}^\text{t}_1, \mathbf{c}^\text{t}_2, \dots, \mathbf{c}^\text{t}_K\}$. The matrix $\mathbf{H}^\text{t}\in\mathbb{F}^{K \times N^\text{t}}$ is constructed, which calculates the Euclidean distance between the target domain clusters and source domain samples $\mathbf{x}^\text{s}$ using Eq.~\ref{eq:src_dis}, and pick the closest samples using Eq.~\ref{eq:min_src_dis}, which provide $Z^\text{t}=\{{z}_1, {z}_2, \dots, {z}_{N^\text{t}}\}$ from $\mathbf{H}^\text{t}_{k,i}$.
Note that here we do not sort $Z^\text{t}$, as the samples in $\mathbf{\widetilde{S}}_a$ are already sorted according to $\mathbf{K}^\text{s}$, which corresponds to the dense region of the source domain. This helps eliminate outliers and ensures that we focus only on the most relevant part of the source.

Afterward, the top $n$ distances from $Z^\text{t}$ are selected and added to $E$. The updated distances are stored in $E= E \cup Z^\text{t}_{1:n}$. We now sort the distances in ascending order $E^*=Sort_{Asc}(E)$ while adding the relevant samples in $\mathbf{R}^*= \mathbf{R} \cup \mathbf{x}^\text{s}_{1:n}$.
Based on ${E}^*$, we reorder all the samples $\mathbf{R}^*=\{\mathbf{x}^\text{s}_1, \mathbf{x}^\text{s}_2, \dots, \mathbf{x}^\text{s}_{m+n}\}$, where $m$ is the total number of existing data, $n$ is the newly added samples. We then select the top $N^\text{r}$ labeled examples:

\begin{equation}
\label{eq:updated_relv}
\begin{aligned}
\mathbf{R}^*=\{\mathbf{x}^\text{r}_i ,y^\text{r}_i\}^{N^\text{r}}_{i=1}
\end{aligned}
\end{equation}

Thus, we estimate the loss of the replay-relevant domain as follows.
\begin{equation}
\label{eq:ce_prev}
\begin{aligned}
\mathcal{L}^\text{r}=-\frac{1}{N^\text{r}}\sum_{i=1}^{N^\text{r}}{y}_i^\text{r} \cdot\log (C(\mathbf{x}^\text{r}_i))
\end{aligned}
\end{equation}

\noindent where $(\mathbf{x}^\text{r}, y^\text{r})$, belongs to the updated replay domain $\mathbf{R}$, that re-calibrated after every $\mathcal{(\widetilde{S}},\mathbf{T)}$ adaptation. Note that $\mathbf{R}$ is a dynamic domain that continues to update with the subset of source domains $\widetilde{\mathcal{S}}$ \footnote{The Algorithm is provided in Section I-B of the supplementary material}. 

\noindent\textbf{Pseudo-label for Target Domain.}
To calculate the pseudo-labels for the target domain, motivated by the augmented confident pseudo-label (ACPL) technique presented in \cite{zeeshan2024subject}, we calculate the target labels by generating an augmented version $\widehat{\mathbf{x}}^t$ of each target sample $\mathbf{x}^t$ using a model $\Theta$. We then estimate the probabilities as $\mathbf{p}^\text{t}=\mbox{softmax}(\mathbf{x}^t)$ and $\widehat{\mathbf{p}}^\text{t}=\mbox{softmax}(\widehat{\mathbf{x}}^t)$, taking the average of two probabilities $a^t=(\widehat{\mathbf{p}}^\text{t}+\mathbf{p}^\text{t})/2$. The selection criteria to assign the pseudo-label to the target sample is determined by the confidence threshold $\tau$ based on $\tau = \tau_0 - \delta \left\lfloor\frac{e}{N}\right\rfloor$, where $e$ the current epoch, $N$ represents the epoch after which $\tau$ decreases, $\delta$ is the reduction value, and $\lfloor \cdot \rfloor$ is the floor function that rounded down to the nearest value. We assign the pseudo-labels to the target samples $\mathbf{\widehat{T}}=(\hat{\mathbf{x}}^\text{t},\hat{y}^\text{t})$ if $a^t$ is greater than $\tau$. At each training step, we estimate the target domain loss as:  
\begin{equation}
\label{eq:ce_tar}
\begin{aligned}
\mathcal{L}^\text{t}=-\frac{1}{N^\text{t}}\sum_{i=1}^{N^\text{t}}\widehat{y}_i^\text{t}\cdot \log(C(\widehat{\mathbf{x}}^\text{t}_i))
\end{aligned}
\end{equation}

\noindent\textbf{Domain Alignment.}
We further mitigate the divergence between the domains using  Maximum mean discrepancy (MMD) \cite{sejdinovic2013equivalence}, which estimates the disparity among two distributions in RKHS space. In our problem, we have three subject domains $(\mathcal{\widetilde{S}}_a,\mathbf{\widehat{T}}, \mathbf{R})$, we jointly calculate the pairwise distances between $(\mathcal{\widetilde{S}}_a,\mathbf{\widehat{T}})$ and $(\mathcal{\widetilde{S}}_a,\mathbf{R})$. For every new source, the disparity is calculated between the target domain, source domain, and replay domain which makes sure to eliminate the domain shift among them.

{\small
\begin{equation}
\begin{split}
\mathcal{D}((\widetilde{\mathcal{S}},\mathbf{\widehat{T}}), (\widetilde{\mathcal{S}},\mathbf{R})) = 
 \frac{1}{N^{\text{s}^2}} \sum_{i\neq j}^{N^{\text{s}}} k(\mathbf{x}_i^{\text{s}}, \mathbf{x}_j^\text{s}) + \frac{1}{N^{\text{t}^2}} \sum_{i\neq j}^{N^\text{t}} k(\mathbf{x}_i^{\text{t}}, \mathbf{x}_j^\text{t}) \\- \frac{2}{N^\text{s} N^\text{t}} \sum_{i=1}^{N^\text{s}} \sum_{j=1}^{N^\text{t}} k(\mathbf{x}_i^\text{s}, \mathbf{x}_j^\text{t}) + \lambda\left\{ \frac{1}{N^{\text{s}^2}} \sum_{i\neq j}^{N^{\text{s}}} k(\mathbf{x}_i^{\text{s}}, \mathbf{x}_j^\text{s}) + \frac{1}{N^{\text{r}^2}} \sum_{i\neq j}^{N^\text{r}} k(\mathbf{x}_i^\text{r}, \mathbf{x}_j^\text{r}) \right\} \\ - \lambda\left\{ \frac{2}{N^\text{s} N^\text{r}} \sum_{i=1}^{N^\text{s}} \sum_{j=1}^{N^\text{r}} k(\mathbf{x}_i^\text{s}, \mathbf{x}_j^\text{r})\right\}
\end{split}
\end{equation}
}

\noindent where $k(.,.)$ indicates a Gaussian kernel, while $\lambda$ is the weight of the contribution of the samples from the replay domain. Thus, to reduce the domain disparity, the alignment loss is defined as

\begin{equation}
\label{eq:ce_dis}
\begin{aligned}
   \mathcal{L}^\text{dis} = \sum_{d=1}^{\widetilde{D}}\mathcal{D}((\widetilde{\mathcal{S}}_a,\mathbf{T}), (\widetilde{\mathcal{S}}_a,\mathbf{R}))
\end{aligned}
\end{equation}

\noindent The $\mathcal{L}^\text{dis}$ is calculated for every $a$-th source domain that belongs to $\widetilde{D}$, i.e., only the selected source subjects. The total target adaptation loss is estimated as

\begin{equation}
\label{eq:total_loss}
\begin{aligned}
   \mathcal{L}^\text{total} = \mathcal{L}^\text{s}+\mathcal{L}^\text{t}+\mathcal{L}^\text{r}+\mathcal{L}^\text{dis}
\end{aligned}
\end{equation}

\noindent The final objective of our multi-subject domain adaptation of source selection is to minimize $\mathcal{L}^\text{total}$.

\begin{table*}
\renewcommand{\arraystretch}{1.6}
\centering
\caption{\centering Accuracy of our Subject-based MSDA and state-of-the-art methods on BioVid for ten target subjects with all 77 sources. \textbf{Bold} text shows the highest and \textit{Italic} shows the second best accuracy.}
\label{tab:biovid_res}
\begin{tabular}{c|c|ccccccccccc}
\thickhline
\textbf{Standards} &\textbf{Methods} & \textbf{Sub-1} & \textbf{Sub-2} & \textbf{Sub-3} & \textbf{Sub-4} & \textbf{Sub-5} & \textbf{Sub-6} & \textbf{Sub-7} & \textbf{Sub-8} & \textbf{Sub-9} & \textbf{Sub-10} & \textbf{Avg} \\ \hline \hline

\multirow{4}{*}{\centering Source Combine} 
& Source-only & 0.62 & 0.61 & 0.65 & 0.55 & 0.51 & 0.71 & 0.70 & 0.52 & 0.54 & 0.55 & 0.59 \\ 
& DANN~\cite{ganin2016domain} & 0.73 & 0.57 & 0.69 & 0.73 & 0.68 & 0.79 & 0.55 & 0.61 & 0.60 & 0.55 & 0.65 \\ 
& CDAN~\cite{long2018conditional} & 0.64 & 0.50 & 0.68 & 0.71 & 0.51 & 0.69 & 0.61 & 0.63 & 0.67 & 0.47 & 0.61 \\ 

\multicolumn{1}{l|}{} & Sub-based (UDA)~\cite{zeeshan2024subject} & 0.73 & 0.64 & 0.73 & 0.59 & 0.54 & 0.75 & 0.76 & 0.53 & 0.51 & 0.58 & 0.63 \\ \hline

\multirow{6}{*}{\centering Multi-Source}
& M\textsuperscript{3}SDA~\cite{peng2019moment} & 0.67 & 0.66 & 0.61 & 0.58 & 0.55 & 0.50 & 0.67 & 0.56 & 0.54 & 0.67 & 0.60 \\ 
& CMSDA~\cite{scalbert2021multi} & 0.93 & 0.47 & 0.81 & 0.87 & 0.53 & 0.84 & 0.57 & 0.54 & 0.74 & 0.70 & 0.70 \\ 
& SImpAI~\cite{venkat2020your} & 0.80 & 0.69 & 0.55 & 0.75 & 0.52 & 0.81 & 0.71 & 0.61 & 0.59 & 0.56 & 0.65 \\ 
& Sub-based~\cite{zeeshan2024subject} & 0.93 & 0.69 & 0.84 & 0.66 & 0.60 & 0.76 & 0.84 & 0.55 & 0.62 & 0.66 & 0.71 \\ 
& Sub-based\textsubscript{top-k}~\cite{zeeshan2024subject} & 0.93 & 0.71 & \textbf{0.86} & 0.87 & 0.88 & 0.92 & 0.86 & 0.77 & 0.84 & 0.68 & 0.83 \\ 

& \cellcolor{lightgray!50}P-MSDA (ours) & \cellcolor{lightgray!50}\textbf{0.99} & \cellcolor{lightgray!50}\textbf{0.76} & \cellcolor{lightgray!50}\textbf{0.86} & \cellcolor{lightgray!50}\textbf{0.92} & \cellcolor{lightgray!50}\textbf{0.89} & \cellcolor{lightgray!50}\textbf{0.94} & \cellcolor{lightgray!50}\textbf{0.87} & \cellcolor{lightgray!50}\textbf{0.81} & \cellcolor{lightgray!50}\textbf{0.98} & \cellcolor{lightgray!50}\textbf{0.78} & \cellcolor{lightgray!50}\textbf{0.88} \\ \hline

\multicolumn{1}{l|}{Fully-Supervised} & Oracle & 0.99 & 0.91 & 0.98 & 0.97 & 0.98 & 0.97 & 0.96 & 0.95 & 0.99 & 0.98 & 0.96 \\ \thickhline                                                                 
\end{tabular}
\end{table*}

\section{Experimental methodology}
\label{sec:experiments}
FER datasets such as RAF-DB \cite{li2017reliable} or AffectNet \cite{kollias2019deep} are widely used but contain mixed individuals, making them less suited for subject-based adaptation where subject variability is crucial. To address this, we evaluate our progressive MSDA approach on two benchmark pain recognition datasets—BioVid Part A \cite{walter2013biovid} and UNBC-McMaster \cite{lucey2011painful}—and extend our study to in-the-wild datasets, Aff-Wild2 \cite{kollias2018aff} and Behavioral Ambivalence/Hesitancy (BAH) \cite{gonzalez2025bah}. While BioVid, UNBC, and BAH explicitly provide subject information, Aff-Wild2 is adapted to a subject-based setup by treating each video as a unique subject, following the protocol in \cite{zeeshan2024subject}. Furthermore, we analyze the impact of progressively incorporating relevant source samples versus full source subjects, shedding light on strategies that best enhance performance and generalization in subject-based adaptation.

\subsection{Datasets}
\noindent\textbf{BioVid Heat and Pain (PartA)} \cite{walter2013biovid}
The dataset comprises 87 subjects recorded in a controlled environment, where each subject is categorized into one of five classes: "no pain" and four escalating pain levels labeled PA1 through PA4, representing increasing pain intensity. Previous studies have indicated that the lower pain intensities, particularly in the initial stages, did not elicit noticeable facial activities. It is recommended that the focus be on the "no pain" and the highest pain intensity categories. Our experiments concentrate on two classes: "no pain" and the highest pain level, PA4. Each subject contributes 20 videos per class, lasting 5.5 seconds each. Following the findings in \cite{werner2017analysis}, which noted that PA4 does not display significant facial activity in the first 2 seconds of the video, we exclude frames from the initial 2 seconds. This ensures that only the latter part of the sequence, where the subject's response to pain is more pronounced, is analyzed.

\noindent\textbf{UNBC-McMaster Shoulder Pain} \cite{lucey2011painful} 
comprises 25 subjects and includes 200 video sequences. Pain intensity for each frame is assessed using the PSPI scale \cite{prkachin1992consistency}, which ranges from 0 to 15. Given the substantial imbalance across pain intensity levels, we adopt the quantization strategy used in \cite{rajasekhar2021deep}, where pain intensities are grouped into five discrete levels: 0 (no pain), 1 (intensity 1), 2 (intensity 2), 3 (intensity 3), 4 (intensities 4-5), and 5 (intensities 6-15). We further evaluate P-MSDA on additional in-the-wild affective computing datasets. Aff-Wild2~\cite{kollias2018aff}, where each video is treated as a subject and 10 diverse target videos are selected, with the rest serving as sources. BAH~\cite{gonzalez2025bah} contains self-recorded videos in semi-uncontrolled settings; we use 10 subjects as targets and the remaining as source domains.

\begin{table}
\small
\renewcommand{\arraystretch}{1.2}
\caption{\centering Accuracy on UNBC-McMaster dataset of our method.}
\label{tab:unbc_res}
\begin{tabular}{c|cccccc}
\thickhline
\textbf{Methods} & Sub-1 & Sub-2 & Sub-3 & Sub-4 & Sub-5 & Avg \\ \hline \hline
\begin{tabular}[c]{@{}c@{}}Source-only\end{tabular} & 0.74 & 0.84 & 0.81 & 0.68 & 0.83 & 0.78 \\ 
DANN & 0.81 & 0.91 & 0.85 & 0.69 & 0.91 & 0.83 \\ 
CDAN & 0.81 & 0.90 & 0.80 & 0.70 & 0.90 & 0.82 \\ \begin{tabular}[c]{@{}c@{}}Sub-based\end{tabular} & 0.76 & 0.87 & 0.84 & 0.70 & 0.85 & 0.80 \\ \hline
\begin{tabular}[c]{@{}c@{}}M\textsuperscript{3}SDA\\ CMSDA\\ SImpAI\\ Sub-based\end{tabular} & \begin{tabular}[c]{@{}c@{}}0.78\\ 0.80\\ 0.80\\ 0.81\end{tabular} & \begin{tabular}[c]{@{}c@{}}0.87\\ 0.86\\ 0.88\\ 0.91\end{tabular} & \begin{tabular}[c]{@{}c@{}}0.92\\ 0.83\\ 0.81\\ \textbf{0.94} \end{tabular} & \begin{tabular}[c]{@{}c@{}}0.66\\ 0.71\\ 0.70\\ 0.72\end{tabular} & \begin{tabular}[c]{@{}c@{}}0.81\\ 0.85\\ 0.87\\ 0.92\end{tabular} & \begin{tabular}[c]{@{}c@{}}0.80\\ 0.81\\ 0.81\\ 0.86 \end{tabular} \\
\rowcolor{lightgray!50} P-MSDA & \textbf{0.87} & \textbf{0.93} & \textbf{0.94} & \textbf{0.74} & \textbf{0.94} & \textbf{0.88} \\ \hline
Oracle & 0.96 & 0.98 & 0.97 & 0.94 & 0.97 & 0.96 \\ \thickhline
\end{tabular}
\end{table}

\subsection{Implementation Detail}

In all experiments, we employ the ResNet18 backbone \cite{he2016deep}, which consists of the encoder $F$ and the discriminative component $C$. To adapt $F$ for subject-based MSDA, we follow the same protocol as ~\cite{zeeshan2024subject} to remove the first
ReLU, the MaxPool layers, and the final 2D adaptive average pooling layer. The backbone and classifier are shared across domains, with images resized to 100$\times$100. Models are trained using SGD with a batch size of 16 and a learning rate of $10^{-4}$. Closest source selection uses $\gamma=0.8$ and $top_s=40$ through empirical evaluation; details are provided in Section III-B of the supplementary material. Target pseudo-labels are generated with $\tau=\tau_0-\delta\lfloor e/N \rfloor$, where $\tau_0=0.90$, $N=20$, and $\delta=0.01$. For ACPL, we follow \cite{zeeshan2024subject} using horizontal flips as augmentation. Replay memory is updated after each source with $top_n$ and $N_r$ set to 2000 samples

\subsection{Baseline Methods}
To evaluate the performance of our method, we define subjects as source and target domains. The first experiment was conducted on the BioVid dataset, where 77 subjects were treated as sources and adapted to the remaining ten target subjects. The following experiment is on the UNBC-McMaster dataset, which includes 20 subjects in the source domain adapted to the remaining five subjects in the target domain. To evaluate the efficacy of our model, we further experimented with a cross-dataset with 20 UNBC-McMaster sources adapted to 10 BioVid target subjects. We follow the previous work of Zeeshan et al.~\cite{zeeshan2024subject} to define the MSDA standard for pain recognition. \textbf{Source-combined:} We first define the lower-bound, which is the traditional approach of training a model by using all the sources and testing the target subject. This approach is also known as source-only, as it does not adapt to the target data. The second experiment combines all subjects as before and then adapts to a target subject as in standard UDA. We compare our method with two standard UDA methods, Domain Adversarial Neural Networks (DANN)~\cite{ganin2016domain}, Conditional Adversarial Domain Adaptation (CDAN)~\cite{long2018conditional}, and one Subject-based~\cite{zeeshan2024subject} UDA method. \textbf{Multi-source DA:} We treat each subject as a separate domain while adapting to the target. We compare our method with three standard and one subject-based MSDA approaches: moment matching for multi-source domain adaptation (M\textsuperscript{3}SDA) \cite{peng2019moment}, implicit alignment (SImpAl) \cite{venkat2020your}, contrastive multi-source domain adaptation (CMSDA), and subject-based domain adaptation \cite{zeeshan2024subject}. M\textsuperscript{3}SDA technique was the baseline method for MSDA classification tasks that reduced the discrepancy based on the moment-matching approach between domains. CMSDA and SImpAI methods are based on generating the target PLs. Subject-based DA is the STA technique used in multi-source domain adaptation for pain estimation.

\noindent\textbf{Oracle:} It is the upper bound where we fine-tune the source model by leveraging labels of every target image in a fully supervised manner.


\section{Results and Discussion}
\label{sec:results}
\subsection{Comparison with State-of-the-Art}
\begin{figure}[t!]
\centering
\includegraphics[width=0.98\linewidth]{ 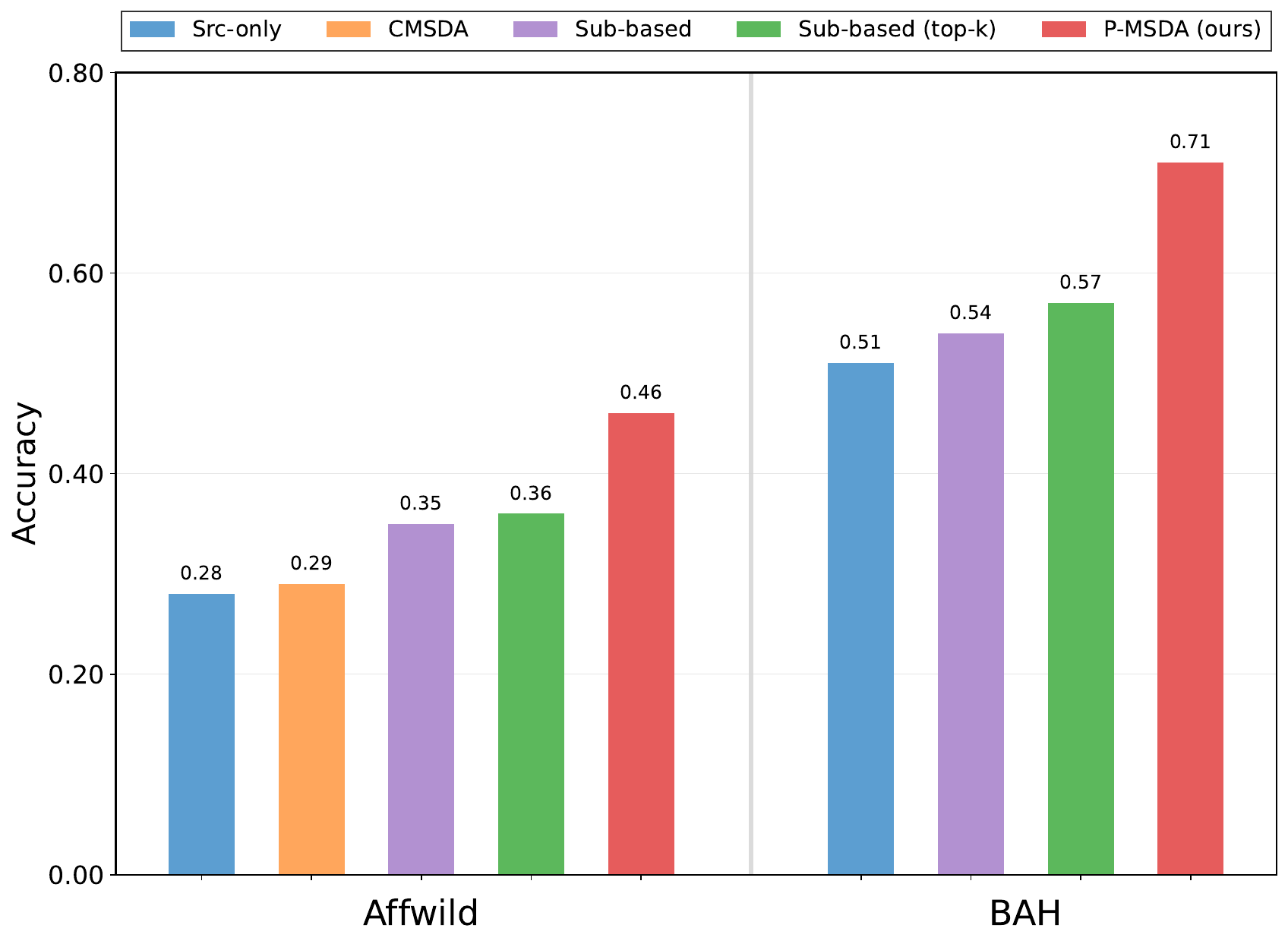}
\caption{Average accuracy on Aff-Wild2 and BAH datasets.}
\label{fig:aff_bah_res}      
\vspace{-10pt}
\end{figure}

The result on the \textbf{BioVid} dataset is shown in Table \ref{tab:biovid_res}. In the source-only method, we follow the lower bound approach without any form of domain adaptation, where a model is trained on training subjects (sources) and evaluated on unlabeled target subjects. In the source-combined (blending) setting, we compare our results with subject-based UDA, DANN, and CDAN. Among these, DANN outperforms the other source-combined methods, achieving an average accuracy of 0.65. We compare our technique with four state-of-the-art MSDA methods. SImpAl, CMSDA, M\textsuperscript{3}SDA, and subject-based MSDA. Our method achieves higher performance for all target subjects with an average accuracy of {0.88} that exceeds the baseline by a large margin. The closest result was with a sub-based\textsubscript{top-k} with an average of {0.83} with a similar performance in \emph{Sub-2} with {0.86} accuracy. Almost every model performance is impressive in subject \emph{Sub-1}. However, our method yields remarkable performance in achieving the accuracy of {0.99}, which is the same as Oracle. The result on \textbf{UNBC-McMaster} is presented in Table \ref{tab:unbc_res}. In source-combined DANN achieves the highest performance of 0.83, outperforming other baseline methods. In contrast, our approach outperforms all methods on every target subject adaptation, including source-only and state-of-the-art MSDA methods, achieving an average accuracy of {0.88}, which is a gain from the previous subject-based method {0.86}. In particular, there is a significant improvement in \emph{Sub-1} performance with an increase of {0.6} compared to the subject-based approach. For \emph{Sub-3}, we match the performance with the subject-based method with {0.94} accuracy.
Fig. \ref{fig:aff_bah_res} reports the average accuracy across 10 subjects from the \textbf{Aff-Wild2} and the \textbf{BAH} datasets. Our proposed P-MSDA achieves the best performance, with 0.46 on Aff-Wild2 and 0.71 on BAH, significantly improving over both source-only and subject-based baselines. These gains highlight the robustness of our approach in handling complex affective states across unconstrained, real-world recordings. A more detailed subject-wise breakdown and additional analysis are provided in the supplementary material.

 \begin{table*}
\renewcommand{\arraystretch}{1.4}
\centering
    \caption{\centering Cross-dataset evaluation: The source model is trained on UNBC-McMaster (20 subjects) and then adapted to 10 BioVid target subjects.}
    \label{tab:cross_unbc_biovid}
\begin{tabular}{c|ccccccccccc}
\thickhline
\textbf{Methods}                                                                                                                                                             & \textbf{Sub-1}                                                         & \textbf{Sub-2}                                                         & \textbf{Sub-3}                                                         & \textbf{Sub-4}                                                         & \textbf{Sub-5}                                                         & \textbf{Sub-6}                                                         & \textbf{Sub-7}                                                         & \textbf{Sub-8}                                                         & \textbf{Sub-9}                                                         & \textbf{Sub-10}                                                        & \textbf{Avg}                                                           \\ \hline \hline
Source-only                                                                                                                                                                  & 0.81                                                                      & 0.54                                                                      & 0.48                                                                      & 0.61                                                                      & 0.52                                                                      & 0.63                                                                      & 0.50                                                                      & 0.48                                                                      & 0.57                                                                      & 0.54                                                                      & 0.56                                                                      \\ \hline 
\begin{tabular}[c]{@{}c@{}}SImpAI~\cite{venkat2020your}\\ CMSDA~\cite{scalbert2021multi}\\ Sub-based~\cite{zeeshan2024subject} \\ Sub-based\textsubscript{top-k}~\cite{zeeshan2024subject}\end{tabular} & \begin{tabular}[c]{@{}c@{}} 0.87\\ 0.88\\ 0.90\\ 0.84\end{tabular} & \begin{tabular}[c]{@{}c@{}} 0.64\\ 0.61\\ 0.50\\ 0.57\end{tabular} & \begin{tabular}[c]{@{}c@{}} 0.67\\ 0.69\\ 0.52\\ 0.50\end{tabular} & \begin{tabular}[c]{@{}c@{}} 0.66\\ 0.55\\ 0.69\\ \textbf{0.85}\end{tabular} & \begin{tabular}[c]{@{}c@{}} 0.58\\ 0.52\\ 0.50\\ 0.72\end{tabular} & \begin{tabular}[c]{@{}c@{}} 0.81\\ 0.79\\ 0.52\\ 0.78\end{tabular} & \begin{tabular}[c]{@{}c@{}} 0.69\\ 0.63\\ 0.53\\ 0.52\end{tabular} & \begin{tabular}[c]{@{}c@{}} 0.53\\ 0.52\\ 0.50\\ 0.52\end{tabular} & \begin{tabular}[c]{@{}c@{}} 0.71\\ 0.77\\ 0.50\\ 0.53\end{tabular} & \begin{tabular}[c]{@{}c@{}} 0.59\\ 0.67\\ 0.58\\ \textbf{0.68}\end{tabular} & \begin{tabular}[c]{@{}c@{}} 0.67\\ 0.66\\ 0.57\\ 0.65\end{tabular} \\
\rowcolor{lightgray!50} P-MSDA (Ours) & \textbf{0.90} & \textbf{0.66} & \textbf{0.80} & \textbf{0.85} &\textbf{0.86}&\textbf{0.86}&\textbf{0.73}&\textbf{0.54}&\textbf{0.94}&\textbf{0.68}&\textbf{0.78}
\\ \thickhline

\end{tabular}
\end{table*}

\subsection{Cross-Dataset Evaluation}
To further evaluate the efficacy of our method, we performed experiments on the cross-dataset setting, where we have 20 UNBC-McMaster labeled source subjects that were adapted to 10 unlabeled target subjects. The result for this setting is shown in Table \ref{tab:cross_unbc_biovid}. We define the baseline as source-only, where the model is trained on source data and evaluated on the target test set. As expected, all the methods outperformed source-only; this is due to the significant domain shift between source and target domains. We further compare our method performance with three different MSDA approaches: SImpAI, CMSDA, and subject-based. Notably, all MSDA approaches demonstrate superior performance compared to the source-only model. This improvement can be attributed to their efficacy in mitigating discrepancies across diverse domains. However, for every target, our method achieves higher performance with an average accuracy of {0.78}, where the subject-based\textsubscript{top-k} matches the performance in \emph{Sub-4} and \emph{Sub-10} with accuracy {0.85} and {0.68}, respectively. It can be observed that MSDA techniques that neglect subject-specific feature representations often fail to optimize the model on the target subject. In contrast, our method demonstrates that choosing the relevant source subject is crucial for effective target adaptation, even when dealing with diversity in the source domain.

\begin{figure}[t!]
\centering
\includegraphics[width=0.98\linewidth]{ 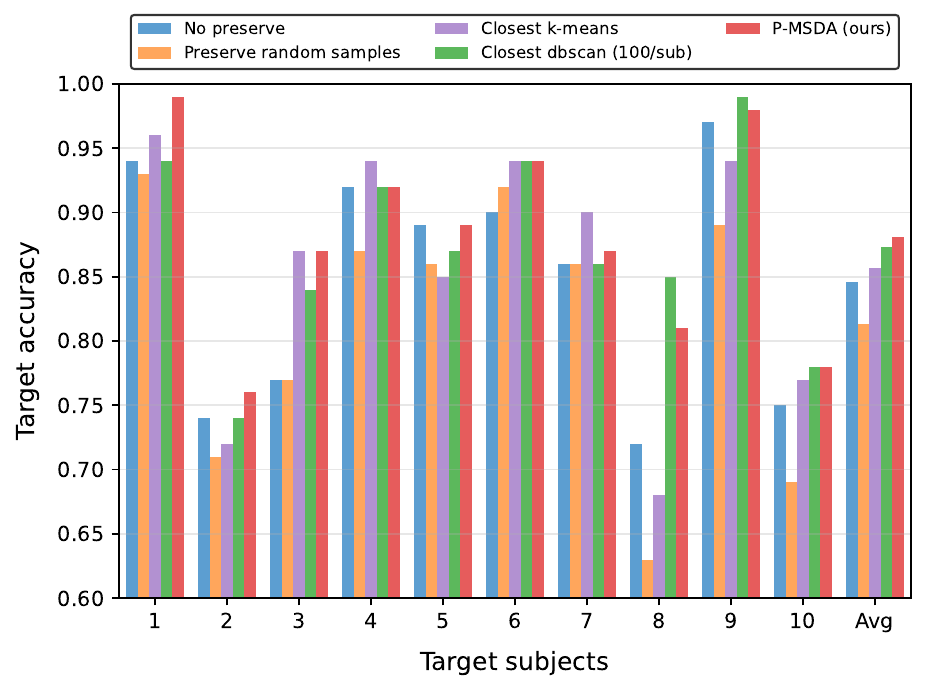}
\vspace{-3pt}
\caption{Comparison of replay sample selection strategies: \emph{No Preserve} (only new subject), \emph{Preserve Random} (fixed random samples), \emph{Closest k-means} (cluster-based), \emph{Closest DBSCAN} (100 samples/subject), and \emph{P-MSDA (Ours)} (density-based pertinent samples).}
\label{fig:abl_relv_sam_graph}      
\vspace{-10pt}
\end{figure}

\begin{figure*}[t!]
\centering
\includegraphics[width=0.98\linewidth]{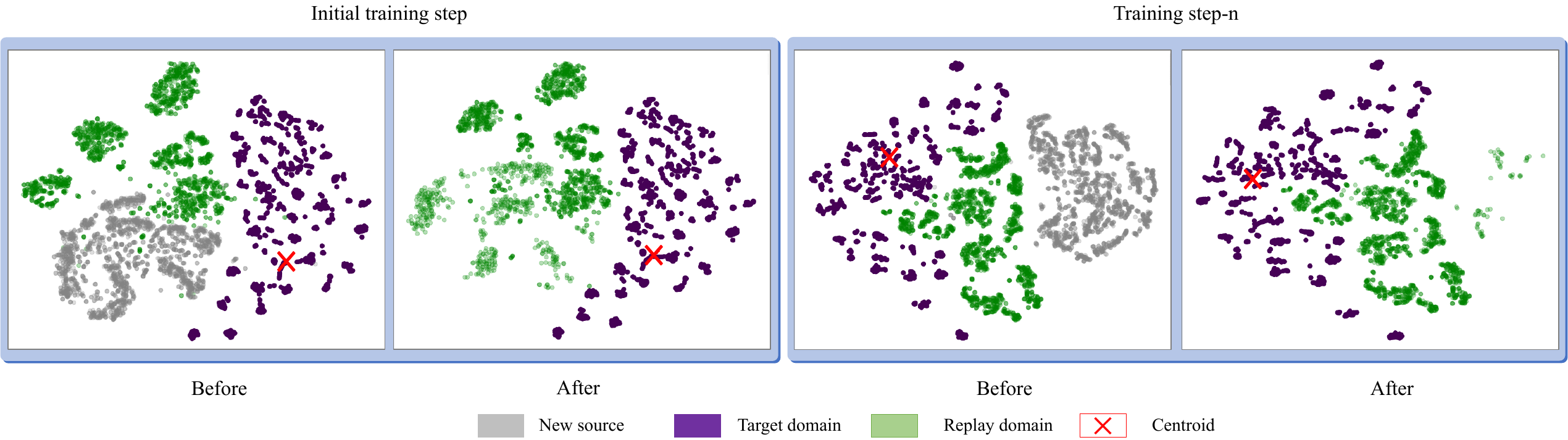}
\caption{T-SNE visualization of Biovid embeddings across source, replay, and target domains (\emph{Subject-1} and \emph{Subject-5}). The replay domain preserves samples over training steps: the \emph{Initial Step} retains more source data, while later steps select fewer but more pertinent samples, reducing distant-subject influence and enhancing target alignment.}
\label{fig:relv_samp_tsne}      
\vspace{-10pt}
\end{figure*}

\begin{figure*}[t!]
\centering
\includegraphics[width=1.0\linewidth]{ sel_closest.pdf}
\caption{Selection of closest source subjects. Example of unlabeled target subject: (a) \emph{Woman 27 (Sub-1)}, (b) \emph{Man 36 (Sub-2)}, (c) \emph{Woman 65 (Sub-9)}, and (d) \emph{Man 25 (Sub-6)} and their respective top-ranked selected source subjects.}
\label{fig:closes_sub_vis}      
\vspace{-10pt}
\end{figure*}

\begin{figure*}[t!]
\centering
\includegraphics[width=0.99\linewidth]{ 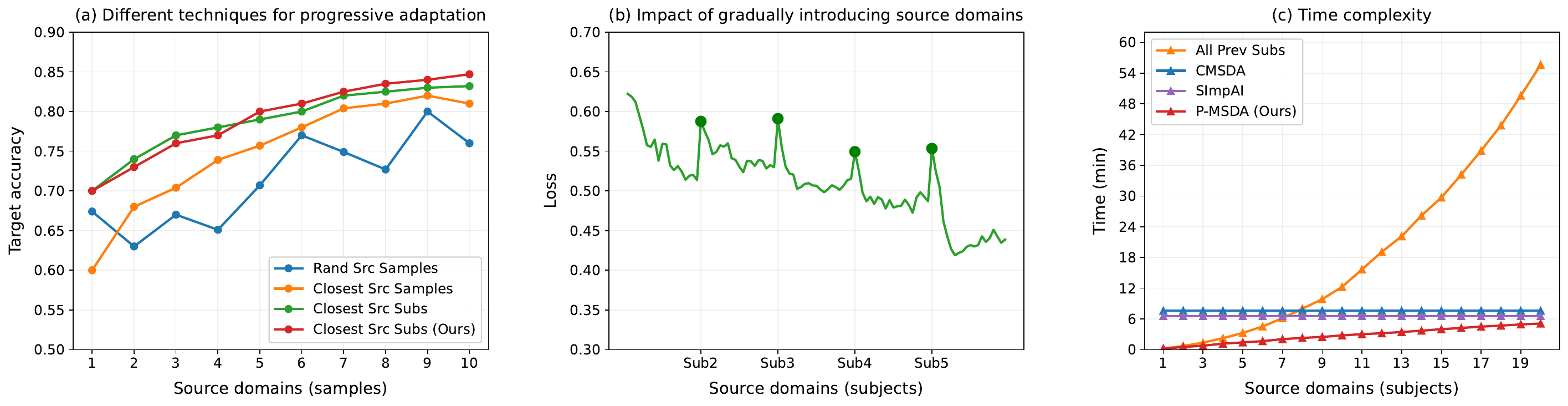}
\caption{(a) We compare four techniques to gradually introduce source data; i) \emph{Random Source Samples}, ii) \emph{Closest Source Samples}, iii) \emph{Closest Source Subjects (all)}, and iv) \emph{Closest Source subjects (ours)}. Every point represents an average of all the target subjects in BioVid. On the x-axis, each increment is equivalent to approximately 2000 samples or a single subject. (b) Loss when introducing a new source domain, adapting to BioVid target subject-10. (c) Time complexity comparison for training; All Previous Subjects, CMSDA, SImpAI, and P-MSDA (ours).}
\label{fig:abl_study}      
\vspace{-10pt}
\end{figure*}





\subsection{Impact of Curriculum-Guided Progressive Adaptation}To evaluate the role of introducing sources in a curriculum-guided manner within our progressive adaptation method, we conducted an ablation comparing ordered adaptation (sorted by similarity) against a non-ordered alternative (shuffled adaptation).
Specifically, we compared four strategies: \emph{(1) Random Source Samples} — subsets of $2000$ images randomly selected from all source domains, introduced progressively to simulate a shuffled or non-ordered adaptation; \emph{(2) Closest Source Samples} — subsets of $2000$ samples selected by cosine similarity from all sources; \emph{(3) Closest Source Subjects (All)} — introducing multiple similar subjects together and retaining all previously seen subjects; and \emph{(4) Closest Source Subjects (Ours)} — progressively introducing entire subjects ranked by similarity to the target while retaining only the most relevant past samples via the replay dictionary.
As shown in Fig. 6(a), the \emph{Random Source Samples} baseline yields the weakest performance, confirming that the absence of ordering is suboptimal. Selecting the \emph{Closest Source Samples} improves over random selection, but still falls short of subject-level ordering, indicating that sample-level similarity alone is insufficient to capture consistent identity-specific cues. While \emph{Closest Source Subjects (All)} initially performs well, the accumulation of too many past subjects reduces focus on newly introduced sources, leading to diminishing gains. In contrast, our curriculum-based \emph{Closest Source Subjects} approach achieves the best and most stable improvement over time, demonstrating that adaptation order is critical. By preserving subject-specific contextual information and gradually increasing domain diversity, this strategy allows the model to learn more coherent and relevant representations, leading to faster convergence and better final accuracy.

\subsection{Impact of Gradually Introducing Source Domains}
We investigate the impact of gradually adding source domains on transfer loss, defined as the discrepancy between source and target domains. This experiment was conducted using the BioVid dataset, with target \emph{Sub-10}. Fig.~\ref{fig:abl_study} (b) illustrates the impact of transfer loss as new sources are introduced. It can be seen that when we added a new source domain, there was a spike in the loss due to the variability between subjects. Following each spike, the model demonstrates its ability to minimize the discrepancy between domains, as evidenced by the subsequent reduction in transfer loss. Furthermore, as training progressed and new subjects were incorporated, we observed a consistent downward trend in transfer loss. This pattern suggests effective model convergence of the selective target subject.

\subsection{Impact of Approaches to Preserve Samples for Replay Domain}

To evaluate the efficacy of our proposed replay strategy, we conducted an ablation study on the BioVid dataset by comparing different techniques for selecting relevant samples, as shown in Fig.~\ref{fig:abl_relv_sam_graph}. We considered five settings: \emph{No Preserve}, \emph{Preserve Random Samples}, \emph{Closest K-means}, \emph{Closest DBSCAN (100/subject)}, and \emph{Ours}. In the \emph{No Preserve} setting, new source subjects are introduced without retaining any previous samples, which results in severe forgetting. In \emph{Preserve Random Samples}, 2,000 samples are randomly selected and retained; however, this performs poorly since random samples fail to capture subject-specific features crucial for adaptation. This highlights the need for a principled sample selection strategy. 
Next, in \emph{Closest K-means}, we apply K-means clustering to identify representative samples. While this improves performance, K-means requires the number of clusters to be specified and is sensitive to outliers, limiting its effectiveness. In contrast, \emph{Closest DBSCAN (100/subject)} uses the DBSCAN clustering algorithm, which does not require a predefined number of clusters and is more robust to outliers, as it relies instead on $\epsilon$ (neighborhood radius) and $minPts$ (minimum samples) to define density and guide cluster formation. This results in a marked improvement over K-means, supporting the use of density-based clustering. Finally, our proposed method extends this idea by applying DBSCAN to build a density-based dictionary of representative samples across subjects, eliminating fixed per-subject constraints and further improving adaptation.

\subsection{Visualization}
\label{sec:relv_sample_tsne} 
\textbf{Density-based Selection of Replay Domain.} Fig.~\ref{fig:relv_samp_tsne} illustrates a t-SNE \cite{van2008visualizing} representation of the embeddings from the last layer of the feature extractor ${F}(.)$ on Biovid target \emph{Subject-1} and \emph{Subject-5}. We show the latent space of the features from a density-based selection of the replay domain before and after selecting a pertinent sample from a new source subject while adapting to a target subject. In the initial training step, the model retains more samples from the newly added source, as the network is more flexible in the initial stage, allowing more relevant samples to be included in a replay dictionary. On the other hand, as training progresses to step n, the replay dictionary is less affected by the newly added source subject. Two notable observations can be made here. 
First, the model gradually acquires useful knowledge from earlier subjects that are more closely related to the target, resulting in fewer samples being drawn from later sources, thereby minimizing the influence of more distant subjects. Second, the samples within the replay dictionary become tightly clustered with the target features, improving alignment with the target domain.
\noindent\textbf{Selection of Relevant Source Domains.} Fig.~\ref{fig:closes_sub_vis} illustrates an example of four target subjects: \emph{Woman 27 (Sub-1)}, \emph{Man 36 (Sub-2)}, \emph{Woman 65 (Sub-9)}, and \emph{Man 25 (Sub-6)} with their respective closest source subjects (using cosine similarity) that are structurally more similar. These subjects are selected first to optimize the model for target adaptation.

\subsection{Complexity Analysis}
\label{sec:train_time}


Fig.~\ref{fig:abl_study} (c) presents the convergence time of the model to the target domain as the number of source domains increases incrementally. This experiment is conducted on the BioVid dataset, using 20 source domains and adapting to the target domain \emph{Sub-4}, with training performed for one epoch on an NVIDIA A100 GPU. 
We evaluate the time complexity across four approaches: \textbf{All Previous Subjects}, \textbf{CMSDA}, \textbf{SImpAI}, and \textbf{P-MSDA (ours)}. In the All Previous Subjects method, all previously visited source domains are retained while progressively adding new ones, making it computationally expensive due to including all subjects at each step. In contrast, CMSDA and SImpAI integrate all source domains at the start of training. As these methods do not involve gradual adaptation, their time complexity is measured after processing all 20 source domains, resulting in a fixed time complexity regardless of the number of included domains. Our proposed approach, P-MSDA, maintains a constant number of three domains: source, target, and replay throughout training. Gradual adaptation is achieved by replacing the current source with a new one and updating the replay domain with previously visited sources. During target adaptation, our method operates with $d$=$3$ domains and approximately $n$=$2000$ samples per domain, resulting in a total sample size $N=n \cdot d$. Training for one epoch takes approximately $t$=$15\cdot d$ seconds. In the All Previous Subjects method, each addition of a new source domain increases the total sample size to $N^*$=$n\cdot (d+1)$, leading to a proportional rise in training time to $t^*$=$t\cdot (d+1)$. However, our method ensures stable training complexity by consistently maintaining $d$=$3$ domains.

\subsection{Limitation of \textbf{P-MSDA}}

\begin{figure}[t!]
\centering
\includegraphics[width=1.0\linewidth]{ 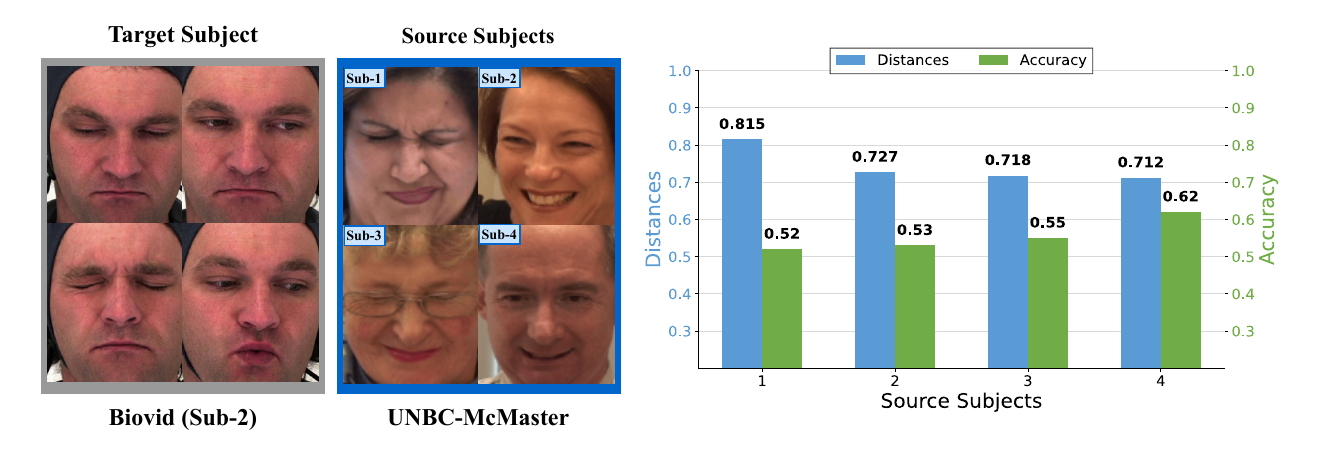}
\vspace{-10pt}
\caption{Adaptation behavior for Sub-2 (young Caucasian male) in cross-dataset. Early adaptation stages select demographically mismatched sources (older Caucasian women) with limited performance gain. In later stages, more structurally relevant sources (older Caucasian males) are selected, leading to a marked accuracy increase.}
\label{fig:limit}      
\vspace{-10pt}
\end{figure}

In the cross-dataset setting, certain target subjects (e.g., Sub-2, Sub-8, and Sub-10) showed lower adaptation performance compared to within-dataset adaptation. This underperformance is mainly due to demographic imbalances and missing expression classes in the source pool, which create a large domain shift between source and target subjects. As shown in Fig.~\ref{fig:limit}, adapting from UNBC-McMaster (source) to BioVid (target) for Sub-2, a young Caucasian male, the first 4 selected sources were older Caucasian women and men. Although closest in learned feature space, these sources offered limited gains due to semantic mismatch. As progressive adaptation began, the model initially struggled to converge due to the large domain shift between source and target, which limited early performance gains. However, as training progressed, the framework gradually reduced this shift, leading to steady improvements over time. This highlights that while P-MSDA may converge more slowly when initial sources introduce a large domain gap, it can still recover as training progresses. Similar cases are observed in the in-the-wild datasets Aff-Wild2 and BAH, where class imbalance and large domain gaps further constrain adaptation. A detailed subject-wise and class-level analysis of these cases is provided in the supplementary material.

\section{Conclusion}
We proposed a novel method of progressively selecting source subjects for personalized facial expression recognition in multi-source domain adaptation. Our model allows the most relevant source subjects to gradually adapt to a target individual in a curriculum manner while preserving the most pertinent samples from the visited sources for the replay domain. We evaluated the efficacy of our model by comparing two scenarios of within and cross-dataset settings. i) for BioVid and UNBC-McMaster, performance is improved significantly for all 10 and 5 target subjects, respectively; ii) For UNBC-McMaster (source) $\rightarrow$ BioVid (target), where subjects are from different domains, our model still outperformed other techniques. Furthermore, we performed a comprehensive analysis of the importance of selecting relevant previous source samples. This enables maintaining important characteristics while adapting to a target subject.

\label{sec:conc}



%



\section*{Acknowledgment}

This research was partially supported by the Natural Sciences and Engineering Research Council of Canada, Fonds de recherche du Québec – Santé, Canada Foundation for Innovation, and the Digital Research Alliance of Canada.

\ifCLASSOPTIONcaptionsoff
  \newpage
\fi

\section{Algorithms}
\label{sec:proposed}
\subsection{Selection of Source Domains}
Algorithm ~\ref{algo:sub_msda_topk} shows the selection of source domains for target adaptation. Initially, we prioritize the source domains with high transferability to align them with the target domain. This will help select source domains with feature distributions similar to the target domain. After aligning the feature distributions of these source domains, a source selection will prioritize the next round of source domains for alignment. As adaptation (training) continues, the model gradually learn to focus on various aspects of the feature distribution to improve transferability. Our approach involves learning a curriculum to prioritize different source domains.

To estimate the similarity matrix for all the source and target domain pairs and selecting the top-N samples,
\begin{equation}
\label{eq:src_sel_cos}
\begin{aligned}
    P =[\mathbf{M}^\text{c}(\mathbf{S}_1, \mathbf{T}),\ldots, \mathbf{M}^\text{c}(\mathbf{S}_D, \mathbf{T})]
\end{aligned}
\end{equation}

\begin{equation}
\label{eq:close_src}
\begin{aligned}
    \widetilde{\mathcal{S}}=\{\mathcal{S}_j: \widetilde{P}_j > \gamma \}  \quad \forall j \in \left\{ 1,\dots ,D\right\}
\end{aligned}
\end{equation}

\begin{algorithm}
\caption{Source Selection for Target Adaptation}
\label{algo:sub_msda_topk}
\begin{algorithmic}[0]
\Require 
\State $\mathcal{{S}}$: set of labeled source domains
\State $\mathbf{T}$: unlabeled target domain
\State $\gamma$: threshold
\State $top_\text{s}$: number of top source domains

\State Initialize: $P \gets []$  \textbf{{\color{gray}$\#$ List to store domain distances}}
\State Initialize: $V \gets []$ 
\textbf{{\color{gray}$\#$ List of adapted (visited) sources}}
\State Initialize $\mathbf{R} \gets \emptyset$ \Comment{$\mathbf{R}$ is a replay domain for $(\mathbf{x}, y)$ pairs}
\end{algorithmic}
\begin{algorithmic} [1]
\While{$|V| < top_\text{s}$}
    \State Filter $\mathcal{{S}}$ to get domains that are not yet adapted
    \For{each domain $\mathbf{S}_a$ in $\mathcal{{S}}$}
        \State \parbox[t]{\dimexpr\linewidth-\algorithmicindent}{
            Compute cosine similarity $p_a$ in mini-batch between samples in $\mathbf{S}_a$ and $\mathbf{T}$ \ref{eq:src_sel_cos} 
        }
        \State Append calculated $p_{a}$ to $P$
    \EndFor

    \State Obtain $\widetilde{P}$ by normalizing $P$ between range $[0-1]$
    \State Select the closest $\widetilde{\mathcal{S}}$ using $\widetilde{P}$ and $\gamma$ by \eqref{eq:close_src}
    \State Append $\widetilde{\mathcal{S}}$ to $V$

    \For{each $\mathbf{\widetilde{S}}_a \in \widetilde{\mathcal{S}}$}
        \State Compute $\mathcal{L}^\text{s}$ for $\mathbf{\widetilde{S}}_a$
        \State Compute $\mathcal{L}^\text{t}$ for $\mathbf{T}$ 
        \State \parbox[t]{\dimexpr\linewidth-\algorithmicindent}{
            \textbf{{\color{gray}$\#$ Initially when $\mathbf{R}$ is $\emptyset$, assign $\mathbf{\widetilde{S}}_a$ }}
        }
        \State    $\mathbf{R} \gets \mathbf{R} = \emptyset \ ?\ \mathbf{\widetilde{S}}_a : \mathbf{R}$  
            \State Compute $\mathcal{L}^\text{r}$ for $\mathbf{R}$

        \State Compute $\mathcal{L}^\text{dis}$ for $\mathbf{\widetilde{S}}_a$, $\mathbf{T}$, and $\mathbf{R}$
        \State \parbox[t]{\dimexpr\linewidth-\algorithmicindent}{

        $\mathbf{R} \gets \textbf{Algorithm~\ref{algo:density_based_samples}}(\mathbf{\widetilde{S}}_a, \mathbf{T}, \mathbf{R})$ \textbf{{\color{gray}$\#$ Update $\mathbf{R}$}}
        
        }
    \EndFor
\EndWhile
\end{algorithmic}
\end{algorithm}

\subsection{Density-based Selection of Replay Domain}
Algorithm ~\ref{algo:density_based_samples} shows our method for \emph{density-based selection of samples in replay domain}. We preserve the relevant samples from the adapted source subject to avoid the forgetting issue while introducing a new source domain into the adaptation process.

Estimate the Euclidean distance between each sample and centroid as,
\begin{equation}
\label{eq:src_dis}
\begin{aligned}
   \mathbf{H}^\text{s}_{j,i} = \left\| \mathbf{x}_i - \mathbf{c}_j \right\|^2 \quad \forall j \in \{1, \dots, K\} ,  i \in \{1, \dots, N^\text{s}\}
\end{aligned}
\end{equation}

Select the closest distance with the cluster centriods and create replay domain,
\begin{equation}
\label{eq:min_src_dis}
\begin{aligned}
   Z^\text{s}=\min_{j \in \{1, \dots, K\}} \{\mathbf{H}^\text{s}_{j,i} \}\quad \forall i \in \{1, \dots, N^\text{s}\}  
\end{aligned}
\end{equation}

\begin{equation}
\label{eq:updated_relv}
\begin{aligned}
\mathbf{R}^*=\{\mathbf{x}^\text{r}_i ,y^\text{r}_i\}^{N^\text{r}}_{i=1}
\end{aligned}
\end{equation}

\begin{algorithm}
\caption{Density-based Selection of Samples in Replay Domain}
\label{algo:density_based_samples}
\begin{algorithmic}[0]
\Require 
\State $\mathbf{\widetilde{S}}_a$: selected source domain
\State $\mathbf{T}$: unlabeled target domain
\State $\mathbf{R}$: replay relevant domain
\Ensure 
Updated replay relevant samples $\mathbf{R}^*$
\State Initialize: $E \gets []$ 
\textbf{{\color{gray}$\#$ List to store distances}}
\State Initialize: $Z^\text{s} \gets []$, $Z^\text{t} \gets []$ 
\textbf{{\color{gray}$\#$ List to store samples}}
\State \parbox[t]{\dimexpr\linewidth-\algorithmicindent}{
    Initialize: Create clusters $\mathbf{K}^\text{s}$ and $\mathbf{K}^\text{t}$ using embeddings from $\mathbf{\widetilde{S}}_a$ and $\mathbf{T}$ to compute centroids $\mathbf{C}^\text{s}$ and $\mathbf{C}^\text{t}$
}
\end{algorithmic}
\begin{algorithmic}[1]
\For{each $\mathbf{x}^\text{s} \in \mathbf{\widetilde{S}}_a$}
    \State Compute $\mathbf{H^s_{k,i}}$ as distances to $\mathbf{C}^\text{s}$ by \eqref{eq:src_dis}
    \State Construct ${Z}^\text{s}$ by selecting the closest samples \eqref{eq:min_src_dis}
\EndFor
\State $\hat{{Z}}^\text{s} \gets Sort_{Asc}({Z}^\text{s})$
\State Reorder $\mathbf{\widetilde{S}}_a$ based on $\hat{{Z}}^\text{s}$

\For{each $\mathbf{x}^\text{s}  \in \mathbf{\widetilde{S}}_a$}
    \State Compute $\mathbf{H^t_{k,i}}$ as distances to $\mathbf{C}^\text{t}$ by \eqref{eq:src_dis}
    \State Construct ${Z}^\text{t}$ by selecting the closest samples \eqref{eq:min_src_dis}
\EndFor

\State Update $E^* \gets E \cup {Z}^\text{t}_{1:n}$ and sort
\State Append top samples to $\mathbf{R}^* \gets \mathbf{R} \cup \mathbf{x}^\text{s}_{1:n}$
\State Select top $N^\text{r}$ examples by \eqref{eq:updated_relv}

\State \Return $\mathbf{R}^*$
\end{algorithmic}
\end{algorithm}

\begin{figure}[t!]
\centering
\includegraphics[width=1.0\linewidth]{ 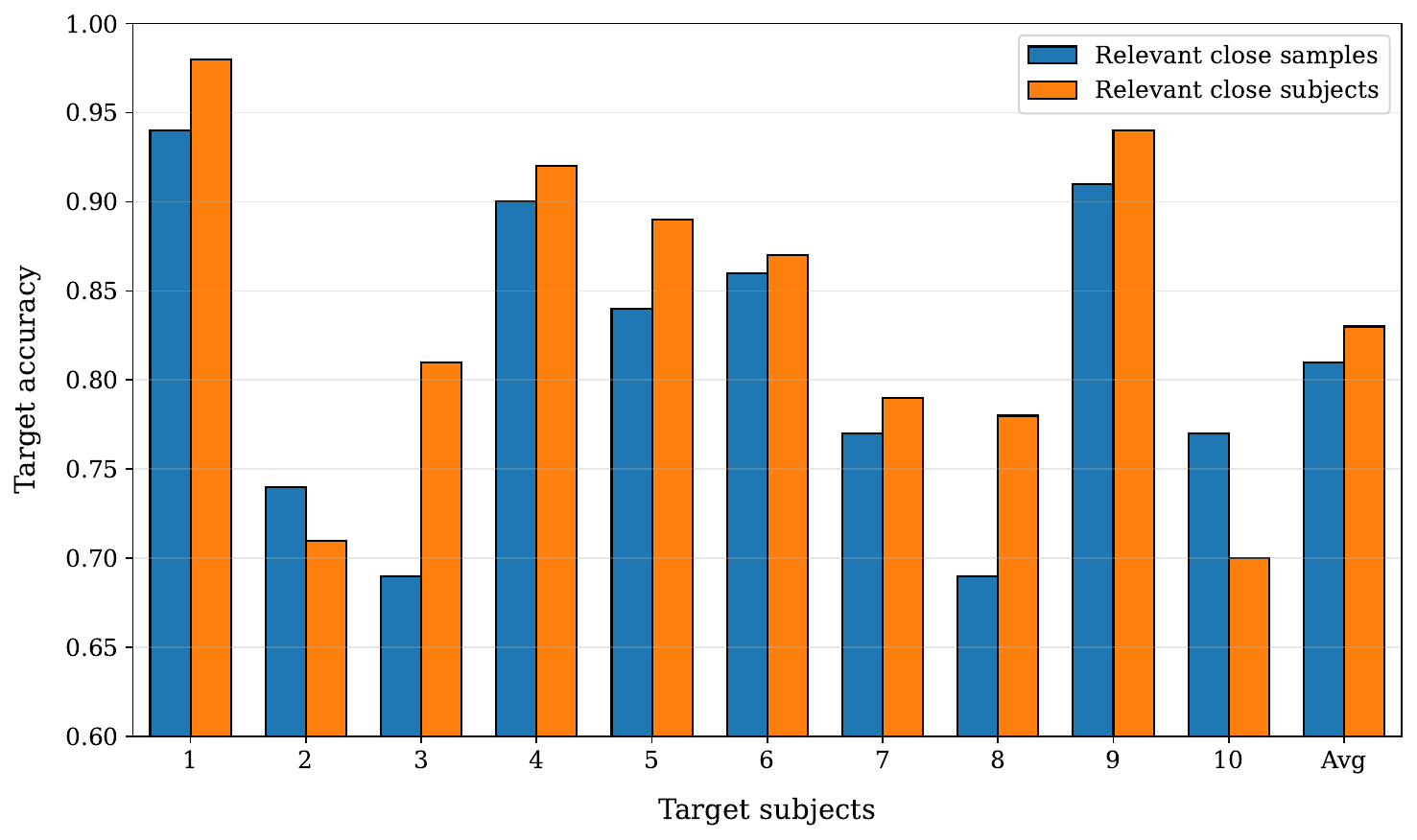}
\caption{Comparison between selecting closest samples versus selecting closest subjects.}
\label{fig:sample_vs_sub_graph}      
\vspace{-10pt}
\end{figure}

\section{Additional Implementation Detail}
\label{sec:experiments}

Our method is implemented on PyTorch \cite{paszke2019pytorch}. For the training of source and target domain subjects, we selected a batch size of 16 with a momentum of 0.9. The learning rate was set to 0.0001, whereas the learning rate between linear and classification layers was set to 0.001. SGD optimizer is chosen with the weight decay of 5e-4. In the training of source subjects, the value of trade-off parameter $\lambda$ is set to 0.1. For the source selection, we set the threshold to \emph{0.90}. The selected source domains are used for the target adaptation, then the next batch of sources were calculated, and the adaptation process will continue until it reaches $top_s$ which is set to $40$ subjects, empirically see section \ref{sec:relv_sample_tsne}. \textbf{Target Subjects:} To make it consistent and comparable across all of our experiments. We select 10 fixed target subject domains from the BioVid Heat Pain Dataset \cite{walter2013biovid}. To have a better target adaptability, we did not select any of the subjects that belong to the 20 unexpressed individuals as cited in \cite{werner2017analysis}. These subjects are:
\verb|081014_w_27|, \verb|101609_m_36|, \verb|112009_w_43|, \verb|091809_w_43|, \verb|071309_w_21|, \verb|073114_m_25|,
\verb|080314_w_25|, \verb|073109_w_28|, \verb|100909_w_65|,     
\verb|081609_w_40|. For UNBC-McMaster we selected 5 target subjects: \verb|107-hs107|, \verb|109-ib109|, \verb|121-vw121|, \verb|123-jh123|, \verb|115-jy115|. For cross-dataset, we use same 10 target subjects from Biovid.

\subsection{Implementation Detail}
In all experiments, we employ the ResNet18 backbone \cite{he2016deep}, which consists of the encoder $F$ and the discriminative component $C$. To adapt $F$ for subject-based MSDA, we follow the same protocol as ~\cite{zeeshan2024subject} to remove the first
ReLU, the MaxPool layers, and the final 2D adaptive average pooling layer. In our experiments, the backbone is shared across every domain, followed by the shared classifier. The images are resized to 100$\times$100 resolution, and the model is trained with stochastic gradient descent (SGD) with a batch size of 16 and a learning rate of $10^{-4}$. For generating target pseudo-labels we set a threshold based on $\theta = \theta_0 - \delta \left\lfloor\frac{e}{N}\right\rfloor$, the initial value of $\theta_0$ is set to \emph{0.91}, that was updated after every $N=20$, with the reduction value $\delta$ is set to \emph{0.01}. For the ACPL technique to generate reliable target PLs, we follow the same setting as Zeeshan et al.~\cite{zeeshan2024subject} and use a horizontal flip as an augmented version of the image. For replay relevant samples, we set $top_n$ and $N_r$ to 2000 samples, which are updated after each newly added source subject.

\begin{table*}[htbp!]
\renewcommand{\arraystretch}{1.4}
\centering
\footnotesize
\caption{{\color{black}\centering Aff-Wild2 accuracy on 10 target subjects for different baselines and our method.}}
\label{tab:affwild_result}
\begin{tabular}{c|ccccccccccc}
\hline\hline
\textbf{Methods} & \textbf{Sub-1} & \textbf{Sub-2} & \textbf{Sub-3} & \textbf{Sub-4} & \textbf{Sub-5} & \textbf{Sub-6} & \textbf{Sub-7} & \textbf{Sub-8} & \textbf{Sub-9} & \textbf{Sub-10} & \textbf{Avg} \\ \hline\hline
Source-only & 0.12 & 0.16 & 0.14 & 0.25 & 0.17 & 0.74 & 0.29 & 0.31 & 0.13 & 0.53 & 0.28 \\ \hline
\begin{tabular}[c]{@{}c@{}}CMSDA~\cite{scalbert2021multi} \\ Sub-based~\cite{zeeshan2024subject}\\ Sub-based\textsubscript{top-k}~\cite{zeeshan2024subject}\end{tabular} &
\begin{tabular}[c]{@{}c@{}} \textbf{0.21} \\ 0.11 \\ 0.12 \end{tabular} &
\begin{tabular}[c]{@{}c@{}} 0.16 \\ 0.17 \\ 0.18 \end{tabular} &
\begin{tabular}[c]{@{}c@{}} 0.18 \\ \textbf{0.20} \\ 0.16 \end{tabular} &
\begin{tabular}[c]{@{}c@{}} 0.20 \\ 0.32 \\ 0.24 \end{tabular} &
\begin{tabular}[c]{@{}c@{}} 0.20 \\ 0.26 \\ 0.20 \end{tabular} &
\begin{tabular}[c]{@{}c@{}} 0.56 \\ 0.85 \\ \textbf{0.88} \end{tabular} &
\begin{tabular}[c]{@{}c@{}} 0.31 \\ 0.27 \\ 0.32 \end{tabular} &
\begin{tabular}[c]{@{}c@{}} 0.52 \\ 0.55 \\ \textbf{0.76} \end{tabular} &
\begin{tabular}[c]{@{}c@{}} \textbf{0.21} \\ 0.15 \\ 0.13 \end{tabular} &
\begin{tabular}[c]{@{}c@{}} 0.37 \\ 0.62 \\ 0.65 \end{tabular} &
\begin{tabular}[c]{@{}c@{}} 0.29 \\ 0.35 \\ 0.36 \end{tabular} \\
\rowcolor{lightgray!50} 
P-MSDA (Ours) & 0.14 & \textbf{0.29} & 0.14 & \textbf{0.57} & \textbf{0.49} & 0.78 & \textbf{0.66} & 0.63 & {0.12} & \textbf{0.76} & \textbf{0.46} \\ 
\hline\hline
\end{tabular}
\end{table*}

\begin{table*}[htbp]
\renewcommand{\arraystretch}{1.4}
\centering
\footnotesize
\caption{\centering \textcolor{black}{BAH accuracy on 10 target subjects with baselines and our method.}}
\label{tab:bah_result}
\begin{tabular}{c|ccccccccccc}
\hline\hline
\textbf{Methods} & \textbf{Sub-1} & \textbf{Sub-2} & \textbf{Sub-3} & \textbf{Sub-4} & \textbf{Sub-5} & \textbf{Sub-6} & \textbf{Sub-7} & \textbf{Sub-8} & \textbf{Sub-9} & \textbf{Sub-10} & \textbf{Avg} \\ \hline\hline
Source-only & 0.34 & 0.59 & 0.32 & 0.71 & 0.67 & 0.37 & 0.44 & 0.51 & 0.70 & 0.47 & 0.51 \\ \hline
\begin{tabular}[c]{@{}c@{}}Sub-based~\cite{zeeshan2024subject} \\ Sub-based\textsubscript{top-k}~\cite{zeeshan2024subject}\end{tabular} &
\begin{tabular}[c]{@{}c@{}} 0.38 \\ 0.38 \end{tabular} &
\begin{tabular}[c]{@{}c@{}} 0.60 \\ 0.60 \end{tabular} &
\begin{tabular}[c]{@{}c@{}} 0.35 \\ 0.49 \end{tabular} &
\begin{tabular}[c]{@{}c@{}} 0.75 \\ 0.76 \end{tabular} &
\begin{tabular}[c]{@{}c@{}} 0.77 \\ 0.78 \end{tabular} &
\begin{tabular}[c]{@{}c@{}} 0.38 \\ 0.46 \end{tabular} &
\begin{tabular}[c]{@{}c@{}} 0.47 \\ 0.45 \end{tabular} &
\begin{tabular}[c]{@{}c@{}} 0.53 \\ 0.51 \end{tabular} &
\begin{tabular}[c]{@{}c@{}} 0.71 \\ \textbf{0.74} \end{tabular} &
\begin{tabular}[c]{@{}c@{}} 0.50 \\ 0.50 \end{tabular} &
\begin{tabular}[c]{@{}c@{}} 0.54 \\ 0.57 \end{tabular} \\
\rowcolor{lightgray!50} 
P-MSDA (Ours) & \textbf{0.67} & \textbf{0.61} & \textbf{0.90} & \textbf{0.89} & \textbf{0.86} & \textbf{0.77} & \textbf{0.64} & \textbf{0.54} & 0.71 & \textbf{0.55} & \textbf{0.71} \\ 
\hline\hline
\end{tabular}
\end{table*}


\begin{table*}[htbp!]
\renewcommand{\arraystretch}{1.4}
\centering
\footnotesize
\caption{{\color{black}Comparison between ResNet18 and ViT backbones on BioVid dataset.}}
\label{tab:biovid_result}
\begin{tabular}{c|c|ccccccccccc}
\thickhline
\textbf{Backbone} & \textbf{Methods} & \textbf{Sub-1} & \textbf{Sub-2} & \textbf{Sub-3} & \textbf{Sub-4} & \textbf{Sub-5} & \textbf{Sub-6} & \textbf{Sub-7} & \textbf{Sub-8} & \textbf{Sub-9} & \textbf{Sub-10} & \textbf{Avg} \\ 
\hline
\multirow{4}{*}{\centering ResNet18} 
& Source-only & 0.62 & 0.61 & 0.65 & 0.55 & 0.51 & 0.71 & 0.70 & 0.52 & 0.54 & 0.55 & 0.59 \\
& Sub-based~\cite{zeeshan2024subject} & 0.93 & 0.69 & 0.84 & 0.66 & 0.60 & 0.76 & 0.84 & 0.55 & 0.62 & 0.66 & 0.71 \\
& Sub-based\textsubscript{top-k}~\cite{zeeshan2024subject} & 0.93 & 0.71 & 0.86 & 0.87 & 0.88 & 0.92 & 0.86 & 0.77 & 0.84 & 0.68 & 0.83 \\
& P-MSDA & \textbf{0.99} & \textbf{0.76} & 0.87 & \textbf{0.92} & \textbf{0.89} & \textbf{0.94} & \textbf{0.87} & \textbf{0.81} & 0.98 & \textbf{0.78} & \textbf{0.88} \\
\hline
\multirow{4}{*}{\centering ViT} 
& Source-only & 0.94 & 0.73 & 0.65 & 0.63 & 0.74 & 0.62 & 0.86 & 0.50 & 0.54 & 0.52 & 0.68 \\
& Sub-based~\cite{zeeshan2024subject} & 0.94 & 0.74 & 0.60 & 0.88 & 0.83 & 0.69 & 0.83 & 0.52 & 0.58 & 0.53 & 0.71 \\
& Sub-based\textsubscript{top-k}~\cite{zeeshan2024subject} & 0.93 & 0.70 & 0.63 & 0.85 & 0.81 & 0.68 & 0.86 & 0.51 & 0.59 & 0.53 & 0.71 \\
& P-MSDA & 0.95 & \textbf{0.76} & \textbf{0.91} & \textbf{0.92} & 0.79 & 0.91 & \textbf{0.87} & 0.78 & \textbf{0.99} & 0.72 & 0.86 \\ 
\thickhline
\end{tabular}
\end{table*}

\begin{table*}[htbp!]
\renewcommand{\arraystretch}{1.4}
\centering
\footnotesize
\caption{{\color{black} Comparison between ResNet18 and ViT backbones on UNBC-McMaster dataset.}}
\label{tab:unbc_result}
\begin{tabular}{c|c|cccccc}
\thickhline
\textbf{Backbone} & \textbf{Methods} & \textbf{Sub-1} & \textbf{Sub-2} & \textbf{Sub-3} & \textbf{Sub-4} & \textbf{Sub-5} & \textbf{Avg} \\ 
\hline
\multirow{3}{*}{\centering ResNet18} 
& Source-only & 0.74 & 0.84 & 0.81 & 0.68 & 0.83 & 0.78 \\
& Sub-based~\cite{zeeshan2024subject}   & 0.81 & 0.91 & \textbf{0.94} & 0.72 & 0.92 & 0.86 \\
& P-MSDA      & \textbf{0.87} & \textbf{0.93} & \textbf{0.94} & \textbf{0.74} & \textbf{0.94} & \textbf{0.88} \\
\hline
\multirow{3}{*}{\centering ViT} 
& Source-only & 0.79 & 0.89 & 0.47 & 0.73 & 0.93 & 0.76 \\
& Sub-based~\cite{zeeshan2024subject}   & 0.78 & 0.88 & 0.80 & 0.70 & 0.91 & 0.81 \\
& P-MSDA      & 0.82 & \textbf{0.93} & 0.90 & 0.71 & \textbf{0.94} & 0.86 \\ 
\thickhline
\end{tabular}
\end{table*}

\begin{table*}[htpb!]
\centering
    \caption{\centering Different source selection criteria. \emph{N-Classes}, determine by training a $N$ source classes; \emph{Maximum Mean Discrepancy (MMD)}, select closest sources from the target; \emph{Cosine-Similarity (CoS)}, pick based on the similarity score.}
    \label{tab:abl_src_sel_tech}
\begin{tabular}{lcccccccccccc}
\toprule
\textbf{Source Selection}  & \textbf{Sub-1} & \textbf{Sub-2} & \textbf{Sub-3} & \textbf{Sub-4} & \textbf{Sub-5} & \textbf{Sub-6} & \textbf{Sub-7} & \textbf{Sub-8} & \textbf{Sub-9} & \textbf{Sub-10} & \textbf{Avg} \\
\midrule
\textbf{N-Classes}               & 0.94           & 0.72           & 0.78           & \textbf{0.92}  & 0.87           & 0.87           & 0.87           & 0.64           & 0.97           & 0.76            & 0.83          \\
\textbf{MMD}                        & 0.97           & 0.75           & 0.83           & \textbf{0.92}  & 0.86           & \textbf{0.94}  & \textbf{0.90}  & \textbf{0.81}  & \textbf{0.98}  & 0.75            & 0.87          \\
\textbf{CoS} & \textbf{0.99}  & \textbf{0.76}  & \textbf{0.86}  & \textbf{0.92}  & \textbf{0.89}  & \textbf{0.94}  & 0.87           & \textbf{0.81}  & \textbf{0.98}  & \textbf{0.78}   & \textbf{0.88} \\
\bottomrule
\end{tabular}
\end{table*}

\begin{table*}[htpb!]
\centering
\caption{Selecting $Top_s$ value for the selection of source domains}
\label{tab:subject_performance}
\begin{tabular}{lccccccccccc}
\toprule
\textbf{\textit{Top\textsubscript{s}}} & \textbf{Sub-1} & \textbf{Sub-2} & \textbf{Sub-3} & \textbf{Sub-4} & \textbf{Sub-5} & \textbf{Sub-6} & \textbf{Sub-7} & \textbf{Sub-8} & \textbf{Sub-9} & \textbf{Sub-10} & \textbf{Avg} \\
\midrule
\textbf{Top-10}            & 0.90 & 0.58 & 0.78 & 0.91 & \textbf{0.89} & 0.91 & 0.84 & 0.74 & 0.76 & 0.65 & 0.796 \\
\textbf{Top-20}            & 0.96 & 0.64 & 0.81 & \textbf{0.92} & 0.83 & 0.92 & \textbf{0.87} & \textbf{0.81} & 0.97 & 0.74 & 0.847 \\
\textbf{Top-30}            & 0.95 & 0.65 & 0.78 & 0.90 & 0.83 & 0.92 & 0.78 & 0.72 & \textbf{0.98} & 0.70 & 0.821 \\
\textbf{Top-40}            & \textbf{0.99} & 0.73 & \textbf{0.87} & 0.92 & 0.88 & \textbf{0.94} & 0.80 & 0.81 & 0.94 & \textbf{0.78} & 0.866 \\
\textbf{Top-50}            & 0.96 & 0.73 & 0.82 & 0.87 & 0.79 & 0.93 & 0.82 & 0.73 & 0.94 & 0.77 & 0.836 \\
\textbf{Top-60}            & 0.95 & 0.72 & 0.61 & 0.90 & 0.81 & 0.90 & 0.79 & 0.61 & 0.85 & 0.71 & 0.785 \\
\textbf{Top-70}            & 0.90 & \textbf{0.76} & 0.81 & 0.91 & 0.49 & 0.85 & 0.80 & 0.68 & 0.51 & 0.72 & 0.743 \\
\textbf{All subjects (77)} & 0.68 & 0.69 & 0.78 & 0.84 & 0.65 & 0.90 & 0.83 & 0.72 & 0.51 & 0.71 & 0.731 \\
\bottomrule
\end{tabular}
\end{table*}

{\color{black}\section{Aditional Results}
\label{sec:results}
\subsection{Results on Additional Datasets}
 We broadened our evaluation beyond pain-related datasets by including two additional datasets from different application domains: (1) Aff-Wild~\cite{kollias2018aff}, a widely used in-the-wild emotion recognition dataset, and (2) the Behavior Ambivalence Hesitancy (BAH) dataset~\cite{gonzalez2025bah}, which captures more naturalistic behaviors under partially uncontrolled conditions. For \textbf{Aff-Wild}, we treated each video as an independent “subject,” excluding any videos containing multiple individuals or duplicate identities to better approximate a subject-specific setting. A total of 10 videos were selected as target subjects, each containing a unique individual. To ensure a diverse emotion distribution, we hand-picked videos with the maximum number of distinct emotion classes (Table~\ref{tab:affwild_all_subs}). The remaining videos served as source subjects. Despite the absence of explicit subject identifiers and the inherent intra-subject variability, P-MSDA achieved an average accuracy of 0.46 (Table~\ref{tab:affwild_result}), outperforming all baselines and demonstrating strong adaptability under challenging in-the-wild conditions. 
 
\noindent For \textbf{BAH}, which provides multiple recordings per subject in semi-uncontrolled environments. We selected 10 target subjects (Table~\ref{tab:bah_targets}): \verb|82711|, \verb|82687|, \verb|82585|, \verb|82592|, \verb|82598|, \verb|82632|,
\verb|82681|, \verb|82683|, \verb|82708|,     
\verb|82714|. The remaining subjects were treated as sources. 

P-MSDA achieved an average performance of 0.71 (Table~\ref{tab:bah_result}), substantially higher than competing methods. These results demonstrate that our approach generalizes to both emotion-oriented FER and behavioral analysis scenarios, and that datasets with explicit subject information allow P-MSDA to fully leverage its personalization strengths. These additional experiments broaden our evaluation beyond pain datasets, demonstrating that P-MSDA generalizes to emotion-oriented FER scenarios and in-the-wild conditions, while also highlighting that datasets with explicit subject identifiers and multiple recordings per subject remain better suited to fully leveraging its personalization capabilities.

\begin{table*}[htbp]
\small
\renewcommand{\arraystretch}{1.2}
\centering
\caption{\centering \textcolor{black}{Class distribution across 10 target subjects of Aff-Wild2. Sub-9 (highlighted) shows a pronounced imbalance.}}
\label{tab:affwild_all_subs}
\begin{tabular}{c|l|c|rrrrrrr|c}
\thickhline
\textbf{Subject} & \textbf{Video-ID} & \textbf{\#Classes} & \textbf{Neutral} & \textbf{Anger} & \textbf{Disgust} & \textbf{Fear} & \textbf{Happiness} & \textbf{Sadness} & \textbf{Surprise} & \textbf{Total} \\
\hline\hline
Sub-1  & 9-15-1920x1080     & 7 & 3483 & 631  & 362 & 1535 & 3019 & 22615 & 2272 & 33917 \\
Sub-2  & 7-60-1920x1080     & 7 & 1914 & 1233 & 2098& 3630 & 4153 & 144   & 1027 & 14199 \\
Sub-3  & video73            & 7 & 139  & 173  & 140 & 127  & 1686 & 156   & 3355 & 5776  \\
Sub-4  & 8-30-1280x720      & 6 & 2204 & 1798 & 716 & 30   & 894  & 0     & 1369 & 7011  \\
Sub-5  & 198                & 6 & 192  & 48   & 121 & 287  & 323  & 0     & 825  & 1796  \\
Sub-6  & 384                & 6 & 1191 & 0    & 27  & 55   & 57   & 20    & 683  & 2033  \\
Sub-7  & video34            & 6 & 2135 & 16   & 99  & 0    & 5895 & 535   & 102  & 8782  \\
Sub-8  & 147                & 5 & 3030 & 0    & 0   & 16   & 81   & 1600  & 16   & 4743  \\
\rowcolor{lightgray!40}
Sub-9  & 121-24-1920x1080   & 5 & 918  & 206  & 0   & 5546 & 281  & 0     & 414  & 7365  \\
Sub-10 & 131                & 5 & 542  & 72   & 62  & 0    & 2362 & 0     & 39   & 3077  \\
\thickhline
\end{tabular}
\end{table*}

\subsection{Comparison between ResNet and ViT Backbones}
We extended our evaluation to include a transformer-based backbone, specifically the Vision Transformer (ViT), alongside the original ResNet18. On the BioVid dataset Table~\ref{tab:biovid_result}, the source-only ViT achieves an average accuracy of 0.68, outperforming ResNet18 0.59, shows that transformers can provide stronger baselines under limited adaptation. However, our proposed P-MSDA substantially boosts ResNet18 to 0.88 and ViT to 0.86, with consistent improvements across subjects, confirming that the method enhances both CNNs and transformer based models. On the UNBC-McMaster dataset Table~\ref{tab:unbc_result}, ResNet18 begins with a stronger source-only baseline (0.78) compared to ViT (0.76), but again P-MSDA improves both backbones, achieving 0.88 with ResNet18 and 0.86 with ViT. These results demonstrate that while ViT can offer competitive baselines, the performance gains from P-MSDA are not dependent on a specific backbone, highlighting that the method is backbone-agnostic and effective for both convolutional and transformer architectures.

}

\section{Ablations}
\label{sec:results}



\subsection{Impact of Selection of Source Domain}
In this ablation, we study different ways of selecting source subjects that are closer to the target domain in the BioVid dataset, shown in Table \ref{tab:abl_src_sel_tech}. In \emph{N-Classes}, we train a ResNet18 model on the 'N' number of classes, where 'N' is the number of source subjects, i.e., 77. The model is trained for 20 epochs, where each class is associated with a source subject. After training, we introduce a target subject to make a prediction. The class with the highest prediction rate will be closest to the target subject. Therefore, we rank the subjects from the highest to the lowest prediction and adapt to the target subject sequentially. This method matches the performance of \emph{Sub-4} with other techniques, achieving an average accuracy of \emph{0.83}. The following criterion was estimating the distance of sources with the target by the maximum mean difference (MMD), improving performance by \emph{0.4}. Finally, when we apply cosine similarity, it exceeds the other two methods by converging to every target subject, significantly improving the adaptation rate. 

\subsection{Choosing Top-s Value for Source Domain Selection}
\label{lab:tops_src_sel}
The closest source selection criterion is based on the $\tau$, which is set to \emph{0.80}, and the $top_s$ is set to $40$ for Biovid subjects (except for sub-2). For UNBC-McMaster, the total number of source subjects is 20, selecting $top_s$ to 20 subjects.
Table \ref{tab:subject_performance} shows empirically the selection criteria of $Top_s$ value. The experiment demonstrates the convergence of the proposed method, with 9 out of 10 target subjects achieving convergence within the first 40 source subjects ($Top_s=40$). \emph{Sub-2} displayed a delayed convergence, which required additional sources to reach stability. This outlier behavior indicates domain-specific characteristics present in the \emph{Sub-2} data distribution, requiring more sources of information for the convergence. 

\begin{table*}[htbp!]
\renewcommand{\arraystretch}{1.2}
\centering
\footnotesize
\begin{minipage}{0.48\textwidth}
\centering
\caption{{\color{black}Effect of varying similarity threshold on accuracy and number of similarity recalculations.}}
\label{tab:similarity_threshold}
\begin{tabular}{c|c|c}
\hline\hline
\textbf{Threshold ($\gamma$)} & \textbf{Accuracy} & \textbf{Re-calc.} \\ \hline\hline
0.50 & 0.93 & 0 \\
0.60 & 0.95 & 1 \\
0.70 & 0.97 & 2 \\
0.80 & \textbf{0.99} & 3 \\
0.90 & 0.95 & 8 \\
1.00 & 0.94 & 40 \\
\hline\hline
\end{tabular}
\end{minipage}
\hfill
\begin{minipage}{0.48\textwidth}
\centering
\caption{{\color{black}Effect of varying pseudo-label threshold ($\tau_{\text{PL}}$) on accuracy.}}
\label{tab:pl_threshold}
\begin{tabular}{c|c}
\hline\hline
\textbf{Threshold ($\tau_0$)} & \textbf{Accuracy} \\ \hline\hline
0.50 & 0.87 \\
0.60 & 0.93 \\
0.70 & 0.95 \\
0.80 & 0.92 \\
0.90 & \textbf{0.99} \\
1.00 & 0.94 \\
\hline\hline
\end{tabular}
\end{minipage}
\end{table*}


\begin{figure*}[htpb!]
\centering
\includegraphics[width=0.98\linewidth]{ 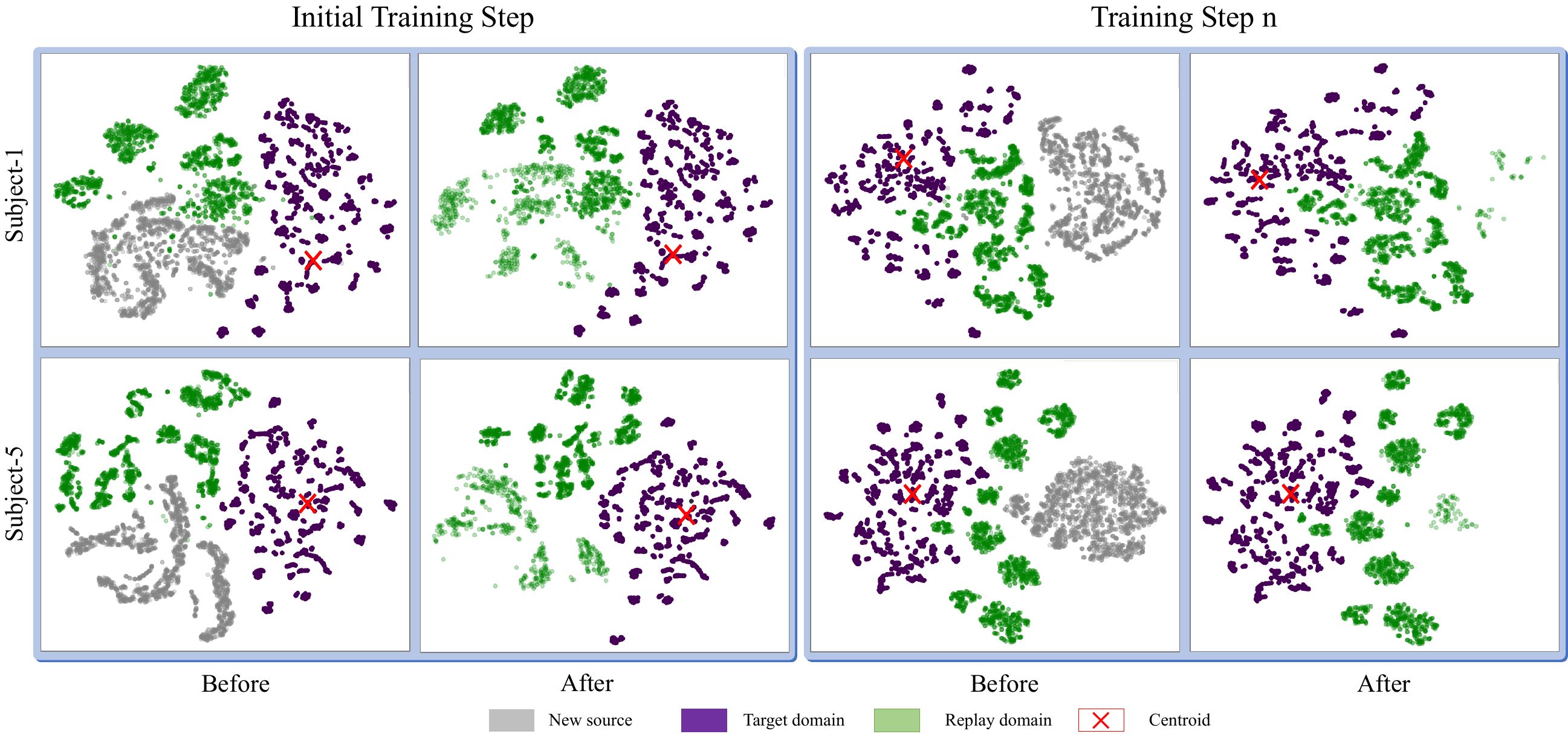}
\caption{T-SNE visualization of patterns from Biovid on source domain, replay domain, and the target domain (\emph{Subject-1} and \emph{Subject-5}). We illustrate how the replay domain preserves samples over different training steps. Before and after, embeddings of selecting pertinent features from the new source subject were extracted. \emph{Initial Step} preserves more samples from the source. \emph{Step n}, incorporate fewer samples from later sources, it minimizes the influence of distant subjects while enhancing alignment with the target.}
\label{fig:relv_samp_tsne}      
\vspace{-10pt}
\end{figure*}

\begin{figure}[htpb!]
\centering
\includegraphics[width=0.98\linewidth]{ 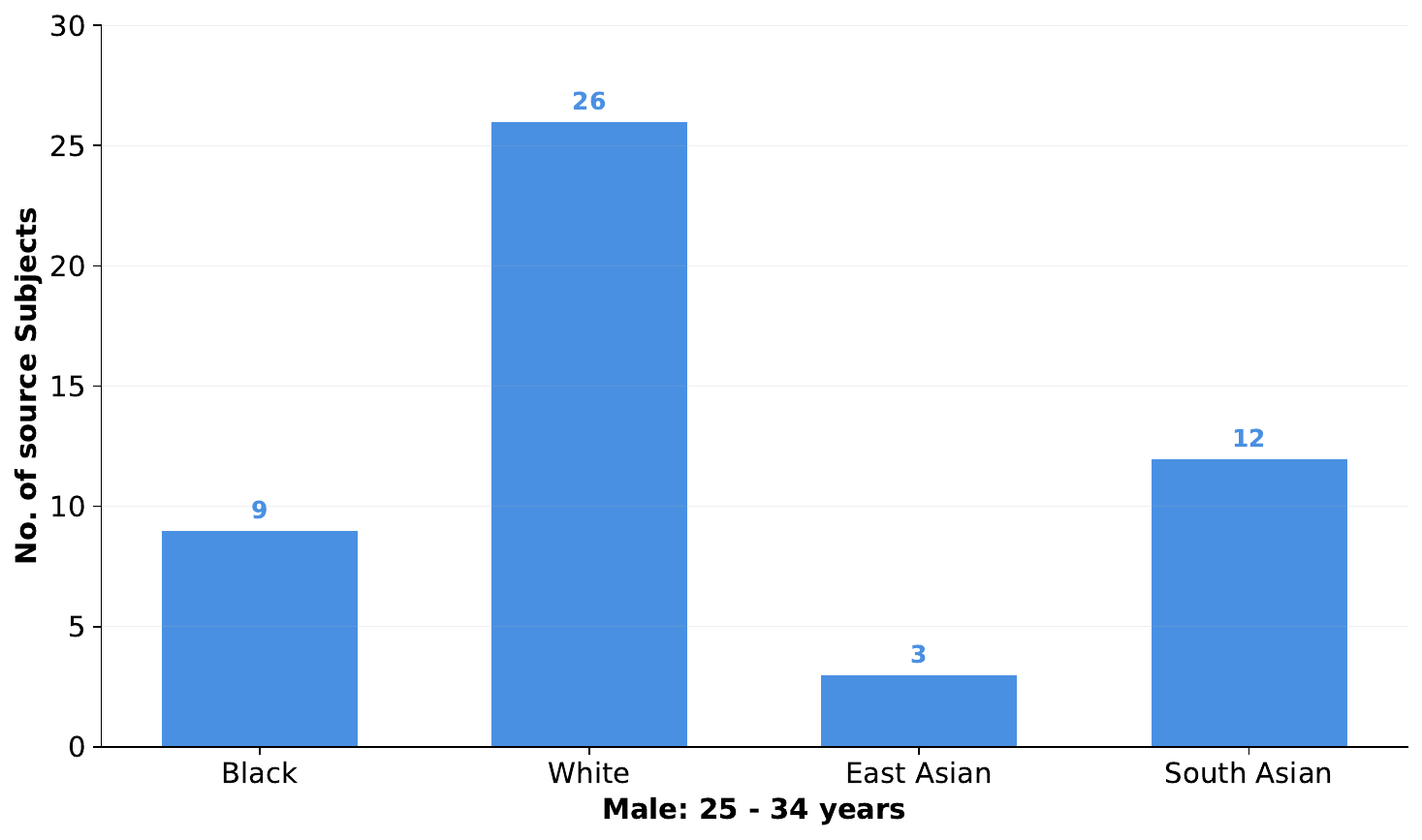}
\caption{\textcolor{black}{BAH source demographic statistics for male, 25-34 age group.}}
\label{fig:bah_src_demo}      
\vspace{-10pt}
\end{figure}

\begin{table*}[htbp]
\renewcommand{\arraystretch}{1.3}
\centering
\footnotesize
\caption{\centering \textcolor{black}{Target subjects of the BAH dataset with demographic and class distributions. N: Neutral, A/H: Ambivalence/Hesitancy.}}
\label{tab:bah_targets}
\begin{tabular}{c|c|c|c|c|c|c|c}
\thickhline
\textbf{Subject} & \textbf{ID} & \textbf{Sex} & \textbf{Age Group} & \textbf{Ethnicity} & \textbf{Frames} & \textbf{A/H Frames} & \textbf{Total} \\ 
\hline \hline
Sub-1  & 82711 & Male   & 25–34 & South Asian     & 2160 & 4376 & 2160 \\
Sub-2  & 82687 & Female & 25–34 & South Asian     & 1550 & 1081 & 1550 \\
Sub-3  & 82585 & Male   & 25–34 & Black           & 2895 & 406  & 2895 \\
Sub-4  & 82592 & Female & 25–34 & Black           & 3685 & 462  & 3685 \\
Sub-5  & 82598 & Female & 25–34 & White           & 5995 & 946  & 5995 \\
Sub-6  & 82632 & Female & 45–54 & White           & 4222 & 1263 & 4222 \\
Sub-7  & 82681 & Female & 25–34 & Middle Eastern  & 1734 & 1068 & 1734 \\
Sub-8  & 82683 & Female & 35–44 & East Asian      & 2165 & 2337 & 2165 \\
\rowcolor{lightgray!50}
Sub-9  & 82708 & Male   & 25–34 & East Asian      & 1229 & 536  & 1229 \\
Sub-10 & 82714 & Male   & 65+   & White           & 1153 & 1169 & 1153 \\
\thickhline
\end{tabular}
\end{table*}

{\color{black}
\subsection{Ablation on Similarity Threshold ($\gamma$).}
For subject selection, we set the similarity threshold ($\gamma$) to $0.80$ in our main experiments. Initially, we select source subjects whose similarity score with the target subject exceeds this threshold. After adapting to the selected sources, we re-calculate similarity scores between the target and all remaining sources until a total of $top_s = 40$ sources is reached, thereby progressively incorporating more closely aligned subjects. Note that an additional ablation on the choice of $N$ sources is provided in Section IV-B of the supplementary material. 
We conducted a study as shown in Table \ref{tab:similarity_threshold} to determine the optimal value of the similarity threshold. If $\gamma$ is set too low, many sources are incorporated at once, which eliminates the opportunity to re-calculate similarities and progressively refine source selection. Conversely, if $\gamma$ is set too high, the system may re-calculate excessively, introducing noise from marginally similar sources. Setting $\gamma = 0.80$ provides the best balance, yielding superior convergence and accuracy by controlling the number of re-calculations while progressively incorporating the most relevant sources.

\subsection{Ablation on Pseudo-Label (PL) Threshold  ($\tau$).} 
Following prior MSDA literature \cite{deng2022robust, scalbert2021multi, zeeshan2024subject}, we chose a relatively high PL threshold to ensure reliable pseudo-label selection and minimize the effect of noisy labels. Moreover, we adopt an adaptive threshold  $\tau = \tau_0 - \delta \left\lfloor\frac{e}{U}\right\rfloor$ to select the value of confidence PL threshold. Specifically, $\tau$ is initialized with $\tau_0 = 0.90$ and is gradually decreased in a step-wise manner, where $e$ denotes the current epoch, $U = 20$ is the update interval, and $\delta$ controls the decrement rate set to $0.01$. This schedule allows the model to start with highly confident pseudo-labels and progressively incorporate more samples as adaptation proceeds.
We further performed an ablation on the initial $\tau_0$ value. As shown in Table \ref{tab:pl_threshold}, lower values (e.g., $\tau_0 \leq 0.80$) significantly degrade performance due to the early inclusion of noisy pseudo-labels before the model is sufficiently adapted, leading to poor convergence. In contrast, a higher starting value (e.g., $\tau_0 = 0.90$) provides the best balance, enabling gradual adaptation while avoiding noise in the early stages. 
}
\subsection{Visualization of Density-based Selection of Relevant Samples}
\label{sec:relv_sample_tsne}
Fig.~\ref{fig:relv_samp_tsne} illustrates a t-SNE \cite{van2008visualizing} representation of the embeddings from the last layer of the feature extractor ${F}(.)$ on Biovid target \emph{Subject-1} and \emph{Subject-5}. We show the latent space of the features from a density-based selection of the replay domain before and after selecting a pertinent sample from a new source subject while adapting to a target subject. In the initial training step, the model retains more samples from the newly added source, as the network is more flexible in the initial stage, allowing more relevant samples to be included in a replay dictionary. On the other hand, as training progresses to step n, the replay dictionary is less affected by the newly added source subject. Two notable observations can be made here. First, the model gradually acquires useful knowledge from earlier subjects that are more closely related to the target, resulting in fewer samples being drawn from later sources, thereby minimizing the influence of more distant subjects. Second, the samples within the replay dictionary become tightly clustered with the target features, improving alignment with the target.
\subsection{\textcolor{black}{Limitation and Failure Cases of P-MSDA}} \textcolor{black}{In addition to the cross-dataset results discussed in the main paper, we also analyzed failure cases in the in-the-wild datasets, Aff-Wild2 and BAH, where the absence of demographically aligned source subjects and highly imbalanced target distributions created significant challenges for P-MSDA. Table~\ref{tab:affwild_all_subs} reports the class distributions for all 10 target subjects of Aff-Wild2, where two types of failure cases emerge. Sub-9 highlights the effect of severe class imbalance, with the Fear class dominating (5,546 samples, ~75\% of the total), while Neutral (918), Anger (206), and Surprise (414) are relatively underrepresented, and both Disgust and Sadness are absent. This extreme skew biases pseudo-labeling and replay toward the majority class, limiting adaptation for minority expressions and contributing to Sub-9 reduced accuracy. In contrast, Sub-1 and Sub-3 represent failure cases caused not by imbalance but by large domain shifts, where no demographically or behaviorally similar source subjects exist. In these cases, the “closest” sources selected early in adaptation are poor matches, constraining transfer and resulting in weaker performance gains. These examples demonstrate that P-MSDA is sensitive to class imbalance (Sub-9) and to domain mismatch (Sub-1, Sub-3), challenges that are common in in-the-wild datasets such as Aff-Wild2.
Similarly, in the BAH dataset, Sub-9 (a male East Asian subject aged 25–34) represents a critical failure case. As shown in Table~\ref{tab:bah_targets}, the target subject belongs to a demographic group that is severely underrepresented in the source pool—only 3 East Asian males (25–34 years) are available compared to 26 White and 12 South Asian males of the same age group (Fig.~\ref{fig:bah_src_demo}). This imbalance constrained the similarity-based selection process, forcing adaptation to rely on less representative sources and producing only modest gains compared to other subjects. The demographic skew of BAH, where certain groups (e.g., White males) are heavily overrepresented while others are nearly absent, amplifies this limitation. These findings highlight that although P-MSDA demonstrates robustness across diverse conditions, its effectiveness diminishes when demographically relevant source subjects are missing, a scenario particularly common in real-world in-the-wild datasets.}



\ifCLASSOPTIONcaptionsoff
  \newpage
\fi



\bibliographystyle{IEEEtran}
\bibliography{main}

\vspace*{-1.1cm}
\begin{IEEEbiography}
[{\includegraphics[width=1in,height=1.25in,clip]{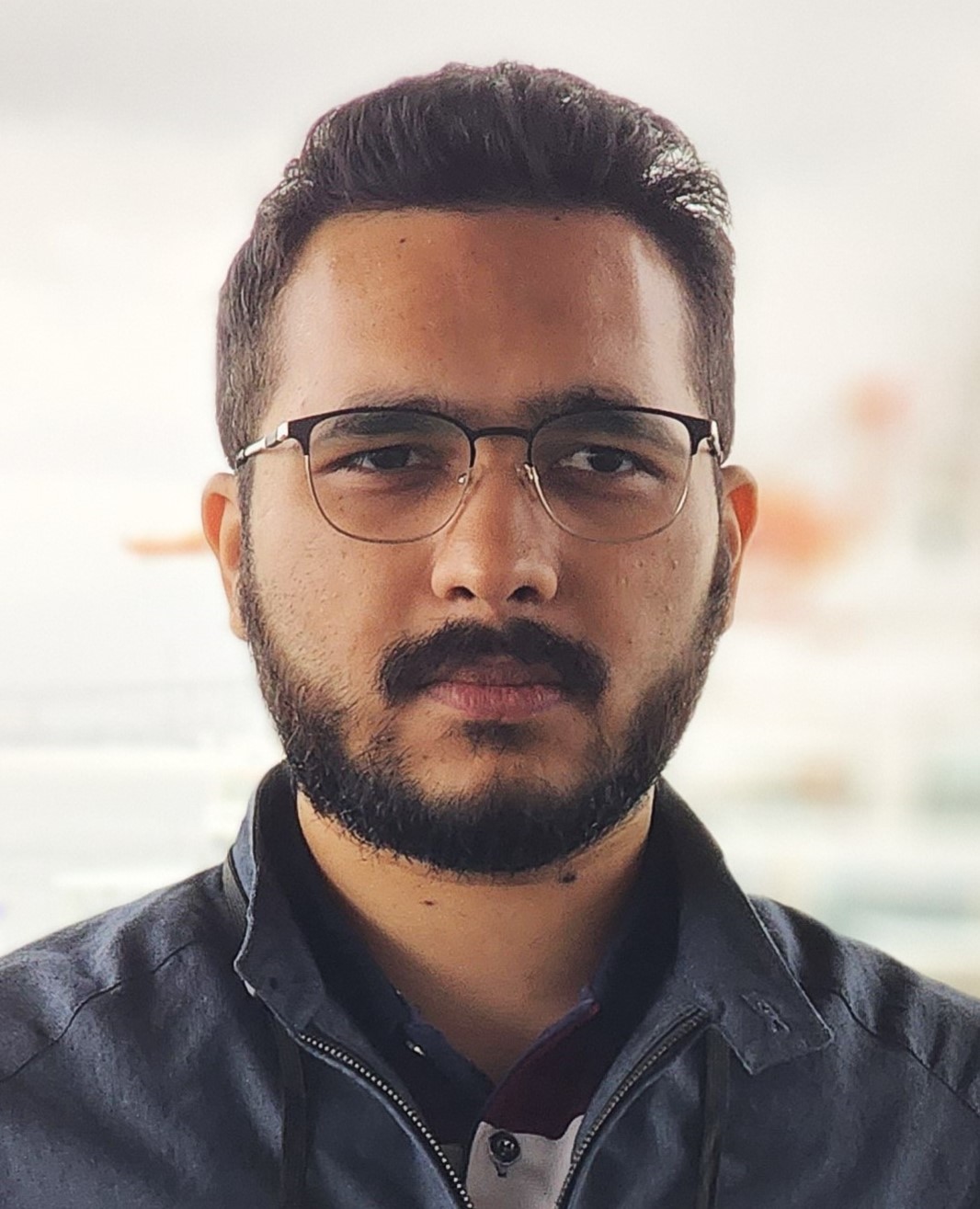}}]{Muhammad Osama Zeeshan} received a Bachelors degree in Computer Sciences from Air University Islamabad and a Masters degree in Data Sciences from Bahria University Islamabad. He is currently pursuing his Doctoral Degree in System Engineering from École de Technologie Supérieure (ÉTS) Montreal, Canada. His research interests include deep learning, computer vision, pattern recognition, unsupervised domain adaptation, and affective computing.

\end{IEEEbiography}

\vspace*{-1.5cm}
\begin{IEEEbiography}
[{\includegraphics[width=1in,height=1.25in]{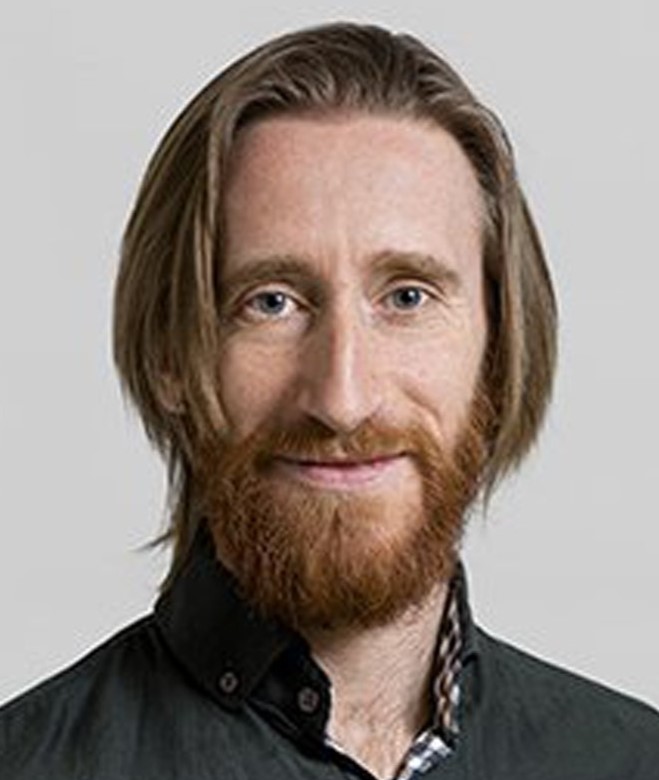}}]{Marco Pedersoli} is associate professor at ETS Montreal. He obtained his PhD in computer science in 2012 at the Autonomous University of Barcelona and the Computer Vision Center of Barcelona. At ETS Montreal he is a member of LIVIA and ILLS and he is co-chairing an industrial Chair on Embedded Neural Networks for Connected Building Control. His research is mostly applied on visual recognition, the automatic interpretation and understanding of images and videos. His specific focus is on reducing the complexity and the amount of annotation required for deep learning algorithms such as convolutional and recurrent neural networks. Prof. Pedersoli has authored more than 50 publications in top-tier international conferences and journals in computer vision and machine learning. 
\end{IEEEbiography}

\vspace*{-1.5cm}
\begin{IEEEbiography}
[{\includegraphics[width=1in,height=1.25in]{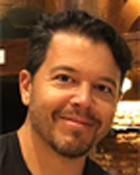}}]{Alessandro Lameiras Koerich} (Member, IEEE) received the Ph.D. degree in engineering from the École de Technologie Supérieure (ÉTS), in 2002. He is currently a professor at the Department of Software and IT Engineering, ÉTS, University of Québec, Montreal. His current research interests include multimodal and trustworthy machine learning and affective computing. He is an associate editor of the IEEE Transactions on Affective Computing, Pattern Recognition, and Expert Systems with Applications. He has served as the general chair of the 14th International Society for Music Information Retrieval Conference, which was held in Curitiba, Brazil, in 2013. In 2004, he was nominated as an IEEE CS Latin America Distinguished Speaker.
\end{IEEEbiography}
\vspace*{-14.9cm}

\begin{IEEEbiography}[{\includegraphics[width=1in,height=1.25in,clip]{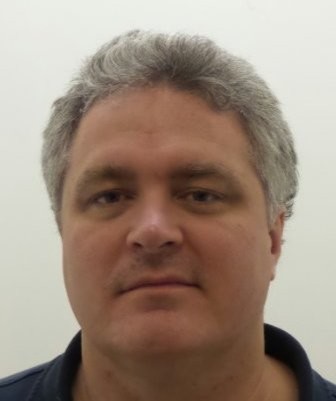}}]{Eric Granger}
(Member, IEEE) received the
Ph.D. degree in EE from École Polytechnique de
Montréal in 2001. He was a Defense Scientist
with DRDC, Ottawa, from 1999 to 2001, and in
Research and Development with Mitel Networks
from 2001 to 2004. He joined the Department
of Systems Engineering, École de technologie
supérieure, Montreal, Canada, in 2004, where he is
currently a Full Professor and the Director of LIVIA,
a research laboratory focused on computer vision
and artificial intelligence. He is the FRQS Co-Chair
in AI and Health, and the ÉTS Industrial Research Co-Chair on embedded neural networks for intelligent connected buildings (Distech Controls
Inc.). His research interests include pattern recognition, machine learning,
and computer vision, with applications in affective computing, biometrics,
face recognition, medical image analysis, and video surveillance.
\end{IEEEbiography}

%



%




\end{document}